\let\mypdfximage\pdfximage
\def\pdfximage{\immediate\mypdfximage}

\documentclass[journal,onecolumn,10pt]{IEEEtran1}

\makeatletter
\def\markboth#1#2{\def\leftmark{\@IEEEcompsoconly{\sffamily}\MakeUppercase{\protect#1}}%
\def\rightmark{\@IEEEcompsoconly{\sffamily}\MakeUppercase{\protect#2}}}
\makeatother

\let\labelindent\relax

\usepackage{times}
\usepackage{marvosym}
\usepackage{longtable}
\usepackage{graphics}
\usepackage{graphicx}
\usepackage{caption}
\usepackage[english]{babel}
\usepackage{subfig}
\usepackage{epsfig}
\usepackage{color}
\usepackage{multirow}
\usepackage{mathrsfs}
\usepackage{textcomp}
\usepackage{amsfonts}
\usepackage{amsmath}
\usepackage{amssymb}
\usepackage{nccmath}
\usepackage[numbers,square,sort&compress,comma]{natbib}
\usepackage{chapterbib}
\usepackage[ruled,vlined,linesnumbered]{algorithm2e}
\usepackage{setspace}
\usepackage{verbatim}
\usepackage{wrapfig}
\usepackage{dblfloatfix}
\usepackage{comment}
\usepackage[strict]{changepage}
\usepackage{ragged2e}
\usepackage{microtype}
\usepackage{enumitem}
\usepackage{nccmath}
\usepackage{hyperref}
\usepackage{doi}
\usepackage{afterpage}
\usepackage{multirow,bigdelim}
\usepackage{booktabs}
\usepackage{color}
\usepackage[outline]{contour}
\usepackage{xcolor}
\usepackage[nameinlink,noabbrev]{cleveref}
\usepackage{bibunits}

\setlist{parsep=0pt,listparindent=\parindent}

{\fontsize{9.75}{10}\selectfont \defaultbibliographystyle{Sledge-JOE-2020-1col-1} \defaultbibliographystyle{IEEEtran}}

\DeclareCaptionFont{singlespacing}{\setstretch{1}}
\numberwithin{figure}{section}
\renewcommand{\thefigure}{\arabic{section}.\arabic{figure}}

\renewcommand \thesection{\arabic{section}}
\numberwithin{equation}{section}

\hypersetup{bookmarks=false,bookmarksopen=false,pdfpagemode=empty,pdfstartview=}

\let\oldnl\nl
\newcommand{\nonl}{\renewcommand{\nl}{\let\nl\oldnl}}

\title{\singlespacing\sf\huge Target Detection and Segmentation in Circular-Scan Synthetic-Aperture-Sonar Images using Semi-Supervised Convolutional Encoder-Decoders}
\author{Isaac J. Sledge, \emph{Member, IEEE}, Matthew S. Emigh, \emph{Member, IEEE},\\ 
Jonathan L. King, \emph{Member, IEEE}, Denton L. Woods, \emph{Member, IEEE},\\
J. Tory Cobb, \emph{Senior Member, IEEE}, and Jos\'{e} C. Pr\'{i}ncipe, \emph{Life Fellow, IEEE}%
\thanks{\fontdimen2\font=1.55pt Isaac J. Sledge is the Senior Scientist for Machine Learning and Dr. Delores M. Etter Emergent Engineer with the Advanced Signal Processing and Automated Target Recognition Branch, Naval Surface Warfare Center, Panama City, FL 32407 (email: isaac.j.sledge.civ@us.navy.mil).  He is the director of the Machine Intelligence Defense (MIND) Lab at the Naval Sea Systems Command.}
\thanks{\fontdimen2\font=1.55pt Matthew S. Emigh is a Research Engineer with the Advanced Signal Processing and Automated Target Recognition Branch, Naval Surface Warfare Center, Panama City, FL 32407 (email: matthew.emigh.civ@us.navy.mil).}
\thanks{\fontdimen2\font=1.55pt Jonathan L. King is an Engineer with the Applied Sensing and Processing Branch, Naval Surface Warfare Center, Panama City, FL 32407 (email: jonathan.l.king.civ@us.navy.mil).}
\thanks{\fontdimen2\font=1.55pt Denton L. Woods is an Engineer with the Littoral Acoustics and Target Physics Branch, Naval Surface Warfare Center, Panama City, FL 32407 (email: denton.woods.civ@us.navy.mil).}
\thanks{\fontdimen2\font=1.55pt J. Tory Cobb is the Senior Technologist for Mine Warfare Automation and Processing with the Advanced Signal Processing and Automated Target Recognition Branch, Naval Surface Warfare Center, Panama City, FL 32407 (email: james.cobb.civ@us.navy.mil).}
\thanks{\fontdimen2\font=1.55pt Jos\'{e} C. Pr\'{i}ncipe is the Don D. and Ruth S. Eckis Chair and Distinguished Professor with both the Department of Electrical and Computer Engineering and the Department of Biomedical Engineering, University of Florida, Gainesville, FL 32611, USA (email: principe@cnel.ufl.edu).  He is the director of the Computational NeuroEngineering Laboratory (CNEL) at the University of Florida.\vspace{0.1cm}}
\thanks{The work of the first and sixth authors were funded by grants N00014-15-1-2013 (Jason Stack), N00014-14-1-0542 (Marc Steinberg), and\newline N00014-19-WX-00636 (Marc Steinberg) from the US Office of Naval Research.  The first author was additionally supported by in-house laboratory independent research (ILIR) grant N00014-19-WX-00687 (Frank Crosby) from the US Office of Naval Research and a Naval Innovation in Science and Engineering (NISE) grant from NAVSEA.  The second, third, and fourth authors were supported by a grant N00014-19-WX-00897 (Dan Cook) from the US Office of Naval Research.  The fifth author was supported via direct funding from the Naval Sea Systems Command.}%
}
\begin{document}
\bstctlcite{IEEEexample:BSTcontrol}

\maketitle
\RaggedRight\parindent=1.5em
\fontdimen2\font=2.1pt
\vspace{-1.55cm}\begin{abstract}\normalsize\singlespacing
\vspace{-0.25cm}{\small{\sf{\textbf{Abstract}}}}---We propose a framework for saliency-based, multi-target detection and segmentation of circular-scan, synthetic-aperture-sonar (CSAS) imagery.  Our framework relies on a multi-branch, convolutional encoder-decoder network ({\sc MB-CEDN}).  The encoder portion of the {\sc MB-CEDN} extracts visual contrast features from CSAS images.  These features are fed into dual decoders that perform pixel-level segmentation to mask targets.  Each decoder provides different perspectives as to what constitutes a salient target.  These opinions are aggregated and cascaded into a deep-parsing network to refine the segmentation.  

We evaluate our framework using real-world CSAS imagery consisting of five broad target classes.  We compare against existing approaches from the computer-vision literature.  We show that our framework outperforms supervised, deep-saliency networks designed for natural imagery.  It greatly outperforms unsupervised saliency approaches developed for natural imagery.  This illustrates that natural-image-based models may need to be altered to be effective for this imaging-sonar modality.\vspace{-1.35cm}
\end{abstract}%
\IEEEpeerreviewmaketitle
\allowdisplaybreaks
\singlespacing

\begin{bibunit}
\bstctlcite{IEEEexample:BSTcontrol}

\vspace{-0.4cm}\subsection*{\small{\sf{\textbf{1.$\;\;\;$Introduction}}}}\addtocounter{section}{1}

The detection of targets via sonar is crucial for mine-countermeasure and unexploded-ordnance-remediation missions, amongst other applications.  Traditional detection approaches rely on processing side-scan, strip-map-based synthetic-aperture-sonar (SSAS) data \cite{HayesMP-jour2009a}.  For this modality, targets of interest are ensonified from a linear path.  A set of pings is recorded that contains backscatter signals from the targets over a limited set of aspect angles.  Such data can be converted into high-resolution imagery.  It can also be represented by acoustic-color plots.  Various machine-learning schemes can then be applied to either representation to extract and analyze potential targets.

Sonar sensing platforms are not limited to just linear survey trajectories when collecting details about underwater scenes.  They can also move in an approximately circular trajectory to ensonify targets of interest \cite{FriedmanAD-conf2005a,CallowHJ-conf2009a,CallowHJ-conf2010a}.  Doing so permits the construction of circular-scan, synthetic-aperture sonar (CSAS) data products \cite{MarstonT-conf2011a,MarstonT-conf2012a,MarstonTM-jour2016a,KennedyJL-jour2014a}.  Due to the use of a circular path, acoustic-color target templates can be obtained, from CSAS data, over the full aspect range.  As well, beamformed CSAS imagery contains details of target and seafloor acoustic reflectivity across multiple aspects.  Such imagery typically provides superior views of scene content compared to SSAS.  It also exhibits greatly improved shape resolvability \cite{MitchellSK-conf2002a,FergusonBG-jour2009a}.  That is, scene geometry is often retained well due to having a larger synthetic aperture than in SSAS imagery.

These traits make CSAS data products well suited for reducing decision-making uncertainty in automated target analyses.  Despite its immense utility, though, CSAS imagery and acoustic-color plots have not yet been used for such applications (see \hyperref[sec2]{section 2}).

Due to the success of deep learning for sonar-based automated target recognition \cite{Valdenegro-ToroM-conf2016a,DenosK-conf2017a,JinL-conf2017a,KohntoppD-conf2017a,WilliamsDP-conf2018a,StewartWK-jour1994a,PerrySW-jour2004a,WilliamsDP-jour2020a,ToppleJM-jour2021a} and other applications \cite{MichalopoulouZH-jour1995a,ChakrabortyB-jour2003a}, we believe that it can be effective for processing CSAS data products (see \hyperref[sec2]{section 2}).  Here, we address the problems of extracting pixel-level target contours and specifying target bounding boxes for CSAS imagery (see \hyperref[sec4]{section 3}).  We focus on detecting small targets, such as debris and ordnance, up to large ones, such as downed aircraft and shipwrecks.

More specifically, we propose a semi-supervised, convolutional encoder-decoder framework for detecting and segmenting targets (see \hyperref[sec4]{section 4}).  We refer to this network as the {\sc MB-CEDN}.  The encoder branch of the {\sc MB-CEDN} extracts contrast features from one or more CSAS images at different spatial scales.  These features are then fed into two decoder branches.  One decoder branch highlights visually salient regions in an unsupervised manner.  The other leverages supervised-trained details to label salient target-like regions.  By salient regions, we mean those that are visually distinct from surrounding ones.  Both candidate segmentation solutions are then aggregated and regularized.  Leveraging solutions from both branches leads to more precise target contours than when using only a single branch's guess.  It also helps when novel target types are encountered.  Regularizing the solutions, by applying random-field-like models, helps to correct localized segmentation mistakes.

We have designed the {\sc MB-CEDN}s to simultaneously process multiple CSAS images.  This functionality is important.  Targets may lose full-aspect coverage, and hence become difficult to spot, in different images of the same general area.  In other images, though, the targets may be observed from multiple aspects.  They may therefore possess strong contrast and anisotropy features that enable them to be reliably segmented.  By warping and combining target segmentations from images of related areas, we can typically overcome a loss in target aspect coverage in one or more CSAS images, thereby improving saliency estimation.  Analogous improvements in classification rates have been demonstrated in \cite{WilliamsDP-conf2016a} when using multiple area-related images.

We evaluate our {\sc MB-CEDN} framework using real-world CSAS imagery.  We illustrate that it reliably delineates multiple targets within one or more CSAS images (see \hyperref[sec5]{section 5}).  We also compare the {\sc MB-CEDN}s against saliency detectors that have been adapted from natural imagery to sonar imagery.  We show that {\sc MB-CEDN}s yield better results than supervised saliency networks and provide justifications for this performance improvement.  {\sc MB-CEDN}s also outperform unsupervised, non-deep saliency segmentation approaches.  Such findings indicate that natural-image approaches require adaptations to handle sonar imagery well.

\phantomsection\label{sec2}
\subsection*{\small{\sf{\textbf{2.$\;\;\;$Literature Review}}}}\addtocounter{section}{1}

A great amount of research has been done for SSAS-based target analysis.  Approaches can be roughly divided into two categories, either supervised or unsupervised.  

Conventional supervised schemes utilize shallow architectures applied to manually extracted features \cite{PettersonMI-conf2005a,DuraE-jour2005a,CartmillJ-jour2009a,MukherjeeK-jour2011a,FandosR-jour2014a,FeiT-jour2015a}.  For example, Dura et al. \cite{DuraE-jour2005a} provide a target detection method that employs kernel-based active learning.  Mukherjee et al. \cite{MukherjeeK-jour2011a} use an image partitioning strategy to segment imagery.  Statistical quantities are extracted from each segmented region for use in a classifier that distinguishes between potential mine and non-mine-like targets.  Cartmill et al. \cite{CartmillJ-jour2009a} applied distributed Bayesian hypothesis testing to locate and classify buried targets.  Deformable and non-deformable template matching has been considered extensively for region segmentation and classification \cite{QuiduI-conf2000a,MignotteM-jour2000b,ReedS-jour2003a}.

Aside from us, Barthomier et al. \cite{BerthomierT-conf2019a} published one of the only articles on deep-network detectors.  They considered generic convolutional networks for creating SSAS-image region proposals.  To date, no supervised, semi-supervised, or unsupervised deep networks have been proposed for target segmentation of SSAS imagery.

For unsupervised detection in SSAS-based imagery, most authors model acoustic highlights and shadows, along with seafloor characteristics, in an attempt to separate targets \cite{AbuA-conf2017a,ChandranV-jour2002a,PezeshkiA-jour2007a,DuraE-jour2008a,TuckerJD-jour2011a,WilliamsDP-jour2015a,FandosR-jour2010a,AbuA-jour2019a}.  Early efforts in this area relied on matched filters \cite{ChandranV-jour2002a} and canonical correlation analysis \cite{PezeshkiA-jour2007a}.  Dura et al. \cite{DuraE-jour2008a} later showed that acoustic shadows could be extracted using Markovian segmentation, which allowed for super-ellipses to be fit to the acoustic highlights to delineate the targets.  Mignotte et al. explored a hierarchical, Markov-random-field approach in \cite{MignotteM-jour2000a}.  Fandos et al. \cite{FandosR-jour2010a} utilized active contour models, among other segmentation schemes, to isolate targets from the seafloor.  Williams \cite{WilliamsDP-jour2015a} developed various efficiently computable confidence measures for target and background characteristics, such as nuisance ripples and shadow details, which were fused to create a target-strength score.  Tucker and Azimi-Sadjadi \cite{TuckerJD-jour2011a} employed a Neyman-Pearson detector for locating targets in dual-band SSAS imagery.

To our knowledge, there has been no published research on CSAS-based target detection and pixel-level segmentation.  Developing this functionality is crucial for many applications.  Both capabilities would enable automated ecological surveys of benthic habitats, like assessing hard coral abundances.  Obtaining accurate coral population estimates would be difficult from detection boxes alone.  Likewise, pixel-level target masks would permit multi-instance recognition of ordnance in acoustic-color plots.  If individual targets are not isolated, then their backscatter responses are merged, which hampers automated analyses.  Other applications, such as cross-domain transfer learning and linguistic change captioning, are also made possible because of pixel-level segmentations.

The aforementioned shallow SSAS target detectors could be applied to CSAS imagery.  Most would be ill suited for it, though.  Many of these detectors rely on acoustic shadows to delineate targets.  Shadows are common in limited-aspect SSAS imagery.  They are often not as conspicuous in multi-aspect CSAS imagery.  SSAS-based detectors are also not designed to natively handle multi-aspect information \cite{CookDA-conf2001a,MidtgaardO-conf2007a}.  Additionally, the manually defined features they use would, potentially, not handle variably sized targets well, let alone multiple target types.

We hence believe that deep-network architectures would be more appropriate for defining robust target detectors.  This is because target representations can be learned in a data-driven fashion.  Those features could then be leveraged to segment the sonar imagery.

In what follows, we review deep networks that could be adapted for sonar target analyses.  We then contrast our network with them to illustrate its novelty.

A majority of target detectors for natural images are convolutional in nature.  Examples include {\sc OverFeat} \cite{SermanetP-conf2014a}, regions-with-CNN-features ({\sc R-CNN}) networks \cite{GirshickR-conf2014a,GirshickR-conf2015a,RenS-coll2015a}, you-only-look-once ({\sc YOLO}) networks \cite{RedmonJ-conf2016a,RedmonJ-conf2017a}, amongst others \cite{HeK-conf2014a,DaiJ-coll2016a}.  These deep networks have exhibited greatly improved performance over classical approaches that rely on either scale-invariant features \cite{LoweDG-conf1999a} or histograms of oriented gradients \cite{DalalN-conf2005a}.  

Object detection and segmentation has been improved, in recent years, by incorporating novel layer types and processing capabilities.  Fully convolutional networks ({\sc FCN}s) \cite{LongJ-conf2015a,DaiJ-coll2016a}, for instance, sum partial scores for each object category across different scales to produce high-quality semantic segmentations.  Hyper-column networks \cite{HariharanB-conf2015a} rely on an analogous approach for object instance segmentation.  Several other schemes like {\sc HyperNet}s \cite{KongT-conf2016a}, inside-outside networks ({\sc ION}s) \cite{BellS-conf2016a}, and {\sc ParseNet}s \cite{LiuW-conf2016a}, concatenate features of multiple layers before computing predictions.  This has been shown to be equivalent to aggregating sets of transformed features.  Single-shot, multi-box detectors ({\sc SSD}s) \cite{LiuW-conf2016b} and multi-scale convolutional networks ({\sc MS-CNN}s) \cite{CaiZ-conf2016a} predict objects at multiple layers of the feature hierarchy without combining features.  

There are also architectures that exploit lateral and skip connections that associate low-level feature maps across resolutions and semantic levels.  Examples include {\sc UNets} \cite{KohlS-coll2018a} and {\sc SharpMask} \cite{PinheiroPO-coll2015a,PinheiroPO-conf2016a}.  Another skip-connection approach was used in an {\sc FCN}-type architecture and combined with Laplacian pyramids to progressively refine the object segmentations \cite{GhiasiG-conf2016a}.  Such skip connections overcome the loss in layer-wise mutual information that occurs after stacking too many layers \cite{YuS-jour2018a,YuS-jour2018b}.  They improve performance even as the network size grows.

A commonality of these approaches is that they are supervised trained.  They require large amounts of labeled data to learn good model parameters.  There is another class of techniques, those that are semi-supervised, that can overcome a lack of data by leveraging weak label annotations.  Papandreou et al. \cite{PanpandreouG-conf2015a} proposed one such approach.  They developed an expectation-maximization-style algorithm for dynamically separating targets.  Souly et al. \cite{SoulyN-conf2017a} advocated using generative adversarial networks ({\sc GAN}s) to generate target examples from a limited set of labeled data.  Wei et al. \cite{WeiY-conf2017a} developed an adversarial erasing approach to localize discriminative image regions for segmenting objects.  More recently, Wei et al. \cite{WeiY-conf2018a} showed that varying convolutional dilation rates can effectively enlarge the receptive fields of convolutional kernels in deep networks.  This property was exploited to transfer surrounding discriminative information to non-discriminative object regions.

Our multi-branch, convolutional encoder-decoder network ({\sc MB-CEDN}) is based on a deep encoder-decoder model with skip connections.  {\sc FCN}-like architectures \cite{LongJ-conf2015a,DaiJ-coll2016a} and {\sc DeconvNet}s \cite{NohH-conf2015a} have a similar composition.  Similar to these networks, the {\sc MB-CEDN} encoder extracts low-dimensional, discriminative image features.  The decoder portion then produces a candidate segmentation from those features. 

There are several differences between {\sc MB-CEDN}s and previous networks.  One is that {\sc MB-CEDN}s features are derived at multiple spatial scales in a manner that is conducive for sonar imagery.  This improves performance for targets of different sizes.  Another difference is that we incorporate two decoder branches, one trained in a supervised fashion and the other in an unsupervised fashion.  The branches produce disparate opinions that can often be combined to improve total performance.  The way that we handle training the unsupervised network branch is also different from previous semi-supervised schemes \cite{PanpandreouG-conf2015a,SoulyN-conf2017a,WeiY-conf2017a,WeiY-conf2018a}.  

Another distinction is that the {\sc MB-CEDN}s can co-segment multiple images.  This functionality is useful when target aspect coverage changes greatly across multiple images of the same general area and hence multiple views of the targets are required to accurately isolate all of the targets in any one image.  

Additionally, we use deep-parsing-network layers to refine the segmentation masks via random-field-like structured prediction.  Instead of applying a random-field-model as a post-processing step, which is traditional, we integrate it with the network.  This has been shown, for natural imagery, to improve performance, since the corrections are learned in conjunction with the classifier, not independently \cite{LiuZ-jour2017a}.  We witness similar behavior in our experiments.  Deep-parsing networks are also more computationally efficient than conventional random-field models.

\phantomsection\label{sec3}
\subsection*{\small{\sf{\textbf{3.$\;\;\;$Circular-Scan Synthetic-Aperture Sonar}}}}\addtocounter{section}{1}

The data we use for model training and testing were collected using a high-resolution, multi-element synthetic-aperture-sonar (SAS) sensor with dual bands.  This sensor has an upper-end center frequency in the hundreds of kilohertz.  The spatial resolution of this band is in the centimeter range.  A low-frequency band is also available, but data from it are not used in this paper.

The SAS sensor was mounted on a Hydroid REMUS 600 underwater vehicle.  This vehicle operated in a variety of littoral and oceanic environments throughout the continental United States.  Here, we consider 2007 underwater scenes.  For many of the scenes, the vehicle maneuvered in a strip-map-search mode so that either human operators or target-detection models could identify potential targets and specify a pivot location.

The vehicle used a circular search pattern to ensonify potential targets from many aspect angles.  Up to three circular passes were made for each scene, with an average radius of 30 m.  Each pass had slightly different center points due to a mixture of positioning-system-estimation errors and vehicle shift caused by water currents.  This center-point shift offered differing views of the scene, changing the target aspect coverage.  The vehicle depth varied for each scene.  Typically, it was at least 5 m above a target.

Using redundant circular trajectories leads to survey times that are often much higher than that of strip-map patterns.  It is common to see a time increase of a factor of two to three.  This higher survey time is offset by the improvement in shape resolvability and hence, typically, an improvement in target saliency.

The backscattered echoes collected by the SAS array were coherently summed.  Vehicle motion compensation and correction, beamforming, and image formation were conducted in manner similar to \cite{MarstonT-conf2011a}.  Targets were brought into focus using a correlation-based scheme \cite{MarstonT-conf2012a,MarstonTM-jour2021a}.  Autofocusing counteracts sound-velocity inhomogeneities, like thermoclines or salinity variations, which can cause image blurring.  It also corrects for aberration spatial variance without the need for markers.  Multi-look processing was used to reduce sonar-image speckle \cite{ChenL-jour2020a}.  Speckle is a type of random, multiplicative noise caused by coherent backscattering interference.

There exists a trade-off between aspect angle and spatial resolution.  Narrower sub-apertures permit better localization in aspect.  Each sub-aperture image will have poorer spatial resolution, though.  We struck a balance by considering 100 sub-apertures that were spaced uniformly by 3.6$^\circ$. 

We logarithmically scaled the CSAS images to enhance scene visibility.  We also color-coded the sub-apertures to improve both scene interpretability and target distinguishability.  Examples are provided in \cref{fig:csas-colormap}.  Sub-apertures were mapped to a continuous hue color wheel.  The colors were determined by the direction of ensonifiction, with 0$^\circ$ corresponding to red, 120$^\circ$ to blue, and 240$^\circ$ to green.  The lightness of the color was determined by a reflectivity power mean.  Saturation was dictated by a reflectivity power mean weighted by the sub-aperture center angle.  This representation links the physical characteristics of scattering direction, scattering intensity, and angular anisotropy with hue, variance, and saturation \cite{PlotnickDS-jour2018a}.  Example images and analyses are provided in the online appendix (see \hyperref[secA]{Appendix A}).  While alternate color mappings could be employed, this one has been studied extensively and repeatedly shown to be conducive for the analysis of targets and benthic habitats.

For each CSAS image, we asked multiple experts to separately provide target segmentation masks.  A consensus was then reached as to the final target boundaries.  Alternate data modalities, such as optical imagery, were sometimes used in this process.  Bounding boxes were then derived from the target contours.

Given that some of our CSAS surveys relied on processed SSAS imagery, it may seem as though target detection in the latter modality is redundant.  Due to vehicle drift, though, we do not necessarily know where a potential target may lie in a CSAS image.  Certain target facets may also become better illuminated in the CSAS imagery, leading to significantly different target boundaries compared to those found in the SSAS image.  Lastly, many of our recent surveys do not have corresponding SSAS imagery.  There is thus a need for CSAS-based target-analysis schemes.

\setcounter{figure}{0}

\phantomsection\label{sec4}
\subsection*{\small{\sf{\textbf{4.$\;\;\;$Methodology}}}}\addtocounter{section}{1}

In what follows, we outline our {\sc MB-CEDN}.  We focus first on the encoder followed by the dual decoding branches and subsequent post-processing operations (see \hyperref[sec4.1]{section 4.1}).  For each of these network components, we motivate our design decisions.  We then describe how we train the {\sc MB-CEDN} (see \hyperref[sec4.2]{section 4.2}).

\begin{figure*}
   \hspace{-0.15cm}\includegraphics[]{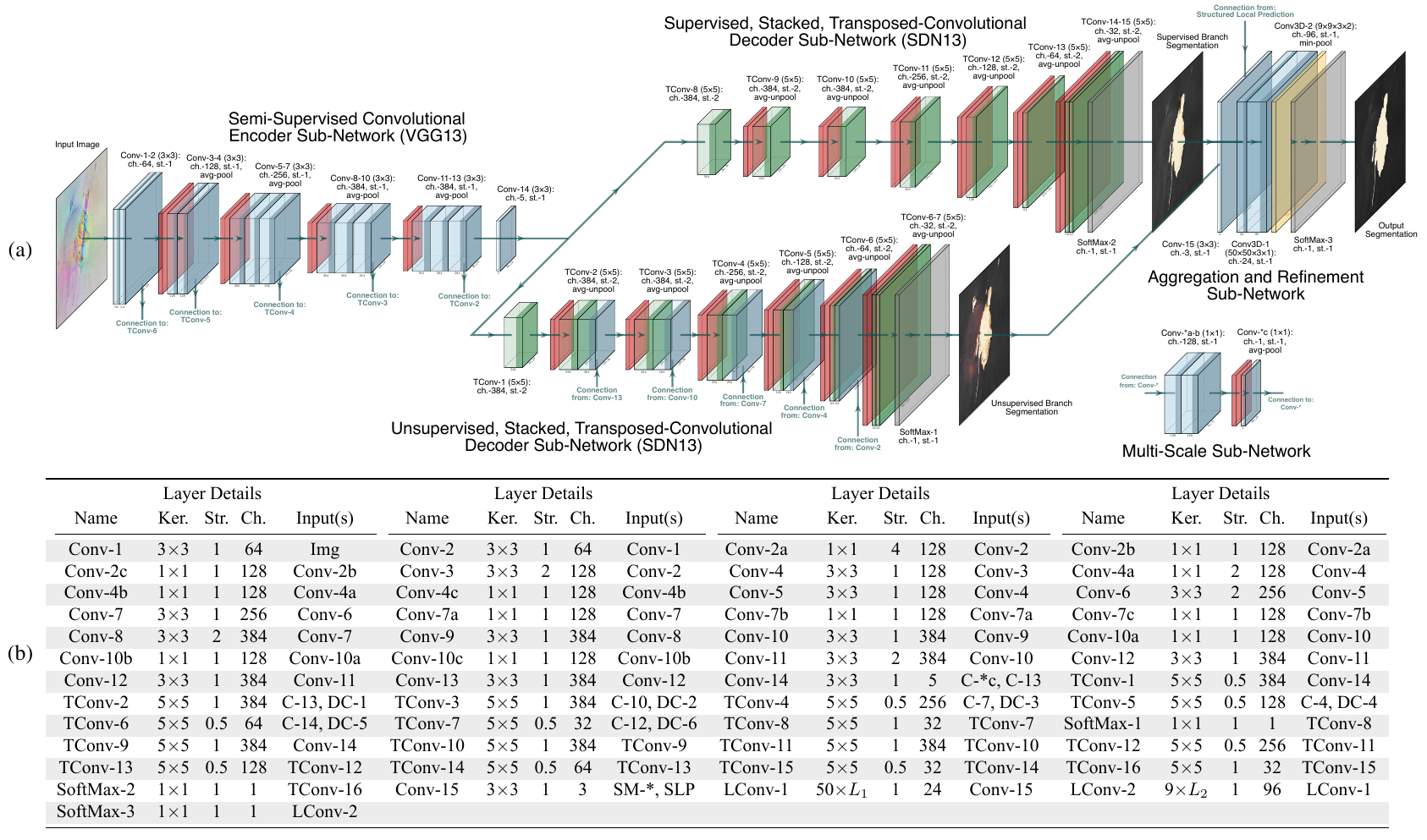}\vspace{-0.15cm}
   \caption[]{\fontdimen2\font=1.55pt\selectfont (a) A network diagram of an MB-CEDN and the resulting segmentation outputs for a CSAS image containing a crashed jet aircraft.  For this diagram, convolutional and transposed-convolutional layers are denoted using, respectively, blue-gray and green blocks.  Average pooling and unpooling layers are denoted using red blocks, while minimum pooling layers are denoted using yellow blocks.  Softmax aggregation layers are denoted using gray blocks.  (b) A tabular summary of the major network layers.  The table is read from left to right and top to bottom.  In some instances, we shorten the names of various layers.  Asterisks are used to indicate that all layers up to that point with a particular naming convention are taken as input.  We recommend that readers consult the electronic version of the paper to see the full image details.\vspace{-0.4cm}}
   \label{fig:mbcedn}
\end{figure*}

\phantomsection\label{sec4.1}
\subsection*{\small{\sf{\textbf{4.1.$\;\;\;$MB-CEDN Network Architecture}}}}

Our framework leverages visual saliency for producing target segmentation masks.  Saliency estimation involves determining what areas in an image are visually conspicuous and hence different from neighboring regions.

For saliency-based target segmentation, visual contrast plays a crucial role, as it affects target perception \cite{ParkhurstD-jour2002a,EinhauserW-jour2003a}.  We therefore have the following design considerations.  First, our {\sc MB-CEDN}s should be deep enough to permit extracting robust contrast attributes from the CSAS imagery that ignore seafloor distractors.  These features should be defined at local and global scales to help detect targets of varying sizes.  Moreover, these features should isolate not only previously seen targets, but also novel ones.

\vspace{0.2cm}\noindent {\small{\sf{\textbf{MB-CEDN Encoder.}}}} To realize these desires, we use a modified {\sc VGG} network ({\sc VGG}13) to extract saliency cues.  These cues include anisotropy and shape features along with contrast details like color, intensity, and orientation.  The {\sc VGG}13 sub-network is shown on the left side of \cref{fig:mbcedn}(a).  The full network specification is given in \cref{fig:mbcedn}(b).

{\sc VGG} networks have been known to provide discriminative features for both natural imagery \cite{Simonyan-conf2015a} and sonar imagery \cite{McKayJ-conf2017a,KvasicI-conf2019a}.  They have been extensively considered in saliency estimation \cite{LiuN-conf2016a,LeeG-conf2016a,LiuN-conf2018a}.  Our use of this network as a feature backbone was motivated by these studies.  Our empirical findings suggest that it is a suitable choice.  Other networks, like {\sc ResNet} \cite{HeK-conf2016a} and {\sc InceptionNet} \cite{SzegedyC-conf2015a}, would likely be worthwhile too.

The {\sc VGG}13 network is composed of five blocks of multiple convolutional layers.  Most blocks are followed by a pooling layer.  Pooling reduces the feature-map size and helps define more semantically rich attributes in deeper blocks.  Leaky rectified-linear-unit ({\sc ReLU}) activation functions are utilized to non-linearly transform the feature maps within each block \cite{NairV-conf2010a}.  As we show in the online appendix, {\sc ReLU}s tend to remove scene distractors, like the seafloor, in inverted-feature reconstructions.  They also emphasize target boundaries well, thereby improving segmentation performance (see \hyperref[secD]{Appendix D}).  Both behaviors likely stem from the sparsity-promoting behaviors of the activation functions \cite{MehtaD-conf2019a}.

As a fully convolutional backbone, {\sc VGG}13 has several advantages.  It can be applied to extract features from arbitrarily sized images.  Non-fully convolutional networks can require fixing image dimensions a priori, which would complicate processing CSAS images if there are major changes in the beamforming and image-formation workflow.  Additionally, fully convolutional networks preserve spatial content.  Some of it is lost during pooling, but this is a necessary compromise to efficiently increase the filter receptive-field size and introduce local transformation invariance \cite{LaptevD-conf2016a}.  Networks with fully connected layers inherently discard spatial coordinates.  This spatial information loss, while sometimes suitable for classification, greatly impedes segmentation (see \hyperref[secB]{Appendix B}).

Conventional {\sc VGG} networks only extract features at a single spatial scale.  To effectively characterize contrast information for differently sized targets, we augment the standard architecture so that it extracts multi-scale contrast features for saliency inference.  We connect three additional convolution layers to each of the first four average-pooling layers.  The first two layers non-linearly transform the feature map.  The third added layer has a single-channel kernel, which is used to predict the initial saliency map.  Although the produced feature maps are of the same size, they are computed using receptive fields with different sizes and hence represent contextual features at different scales.  We further stack these four feature maps with the last output feature map of the {\sc VGG}13 network.

There are additional changes that we make to the {\sc VGG} network.  The standard version consists of five global pooling layers which have an increasingly smaller receptive field containing contextual information about salient targets.  Each pooling layer reduces the feature-map dimensionality by about half.  This leads to low-resolution predictions, which can complicate saliency-map estimation.  The {\sc MB-CEDN} would have to rely on copious amounts of training data to accurate process and progressively upsample the segmentation masks in the decoder branches.  We avoid this issue by increasing the prediction resolution.  We modify the stride for the final two pooling layers, which prevents unnecessary downsampling.  Additionally, to maintain the same receptive-field size in the remaining layers, we apply an \`{a}-trous dilation \cite{ChenLC-jour2018a} to the filter kernels.  This allows for efficiently controlling the resolution of convolutional feature maps without the need to learn extra parameters, thereby improving training times.

\begin{figure*}
   \hspace{-0.15cm}\includegraphics[]{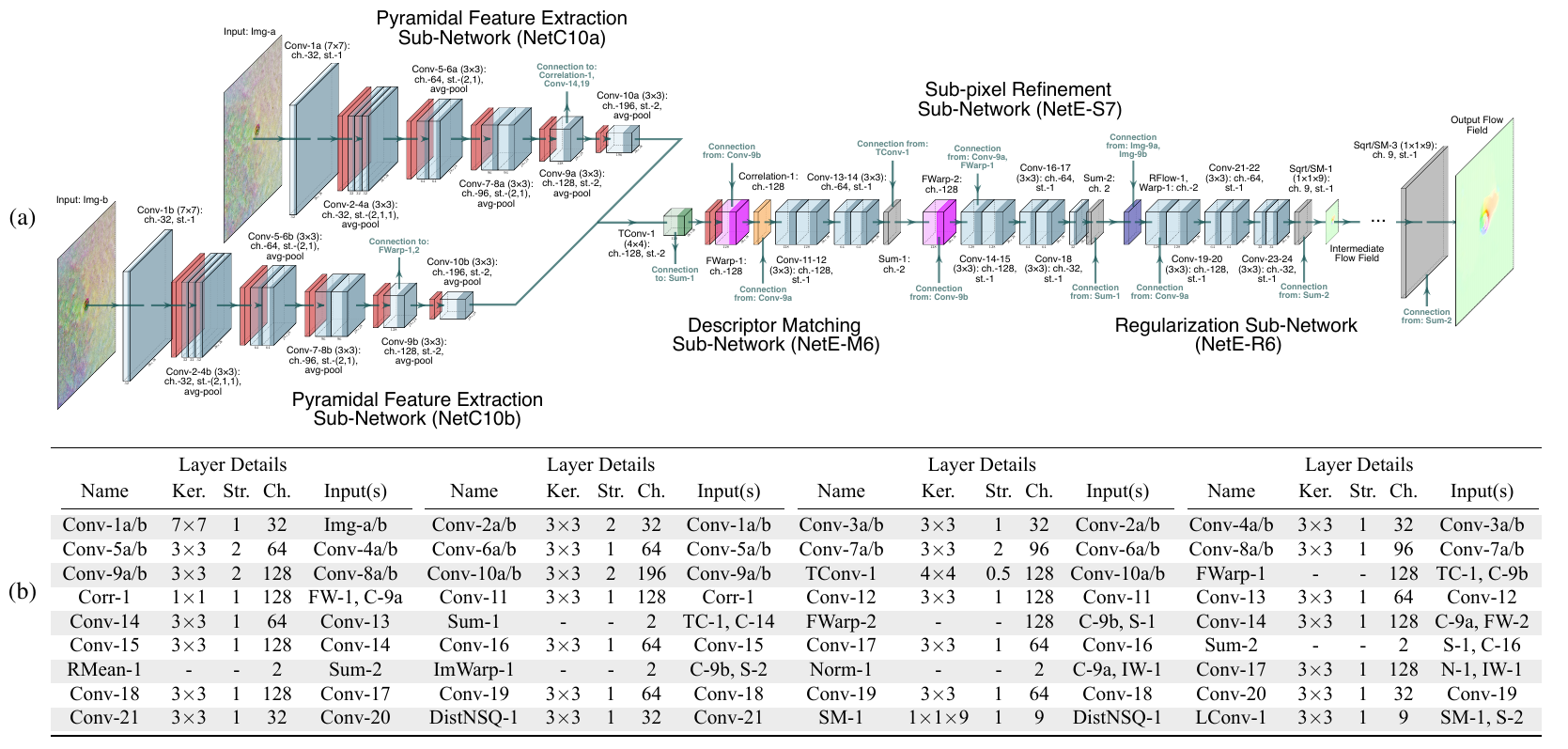}\vspace{-0.15cm}
   \caption[]{\fontdimen2\font=1.55pt\selectfont (a) A network diagram for the small-scale {\sc FlowNet} model, {\sc LiteFlowNet} (LFN), used to estimate the dense, large-displacement scene correspondence between two CSAS images of a clutter target.  For this diagram, correlation layers are denoted using orange blocks.  Flow warping and image warping layers are, respectively, represented by pink and dark blue blocks.  The remaining layers follow the color scheme outlined in \cref{fig:mbcedn}.  (b) A tabular summary of the major network layers.  The table is read from left to right and top to bottom.  In some instances, we shorten the names of various layers.  Asterisks are used to indicate that all layers up to that point with a particular naming convention are taken as input.  We recommend that readers consult the electronic version of the paper to see the full image details.\vspace{-0.4cm}}
   \label{fig:lfn}
\end{figure*}

Another change we make is the pooling process.  VGG13 networks typically rely on global max pooling.  Max pooling \cite{OquabM-conf2015a} has been shown to localize objects in weakly supervised networks, albeit only at a single point versus their full extent.  Using such an operation would help learn features to detect and reliably classify targets but impede easily segmenting them.  Our use of average pooling encourages the network to identify the complete extent of the target \cite{ZhouB-conf2016a}, which aids in segmentation.  This occurs because, when computing the map mean by average pooling, the response is maximized by finding all discriminative parts of an object.  Those regions with low activations are ignored and used to reduce the output dimensionality of the particular map.  For global max pooling, low scores for all response regions, except that which is most discriminative, do not influence the resulting layer output.  Discriminative regions of the target may hence be ignored in the dimensionality reduction process, thereby complicating their localization.  Moreover, global max pooling has the tendency to propagate sonar speckle \cite{WilliamsDP-jour2020a}.  Useful features about the targets can hence be inadvertently removed in the initial layers.

\vspace{0.2cm}\noindent {\small{\sf{\textbf{MB-CEDN Decoders.}}}} The {\sc MB-CEDN} has two stacked decoder sub-networks, as shown on the right side of\\ \noindent \cref{fig:mbcedn}(a).  These convert the bottleneck-layer encoder features into saliency-based segmentation and detection estimates.  As with the encoder, we consider a fully convolutional design for the decoders to preserve spatial information throughout segmentation inference.

The first decoder branch isolates targets in a supervised-trained fashion.  This branch uses a six-block, thirteen-layer topology ({\sc SDN}13) with a series of transposed, \`{a}-trous convolutions.  Global average unpooling layers are inserted to upsample the segmentation estimate so that it is eventually the same size as the input image. 

The second branch of the {\sc MB-CEDN} ({\sc SDN}13) is trained in an unsupervised manner to be a salient target detector.  Its purpose is to highlight all visually salient regions, not just those that may contain targets.  This branch utilizes a similar network topology as the first branch.  It includes additional layers, compared to the supervised branch, which we found helped improve performance.

We avoid using recurrent-convolutional layers, unlike some authors \cite{LiangM-conf2015a,LiuN-conf2016a}, to refine the upsampled saliency maps.  Such layers can be hard to train, since their temporal-sequence memory is tied to the state space size.  They also did not appear to improve results greatly enough to warrant inclusion in our {\sc MB-CEDN}s (see \hyperref[secB]{Appendix B}).

The responses from both branches are converted into probabilities via a softmax layer before being aggregated by structured local predictors \cite{BuloSR-conf2012a}.  These models implement a selective saliency merging process that respects image content.  The result is then cascaded into a final convolutional layer before being passed to the multi-image aggregation and refinement sub-network.

Due to how the branches are trained, the interplay of both leads to a mixture of bottom-up \cite{GaoD-conf2007a,KleinDA-conf2011a,HouX-conf2007a} and top-down saliency \cite{JuddT-conf2009a,JiaY-conf2013a,ShenX-conf2012a,LiuR-conf2014a}.  Top-down saliency is a memory-based, goal-focused process to discern what portions of an image are salient.  It is primarily implemented by the supervised decoder.  Top-down saliency handles previously encountered target types well.  It is also helpful for imaging sonar.  The conventional center-surround assumption \cite{GaoD-conf2007a} found in natural imagery does not always hold for our application.  Due to poor aspect coverage, portions of targets may not be visually conspicuous.  Bottom-up saliency is a memoryless, stimulus-driven process.  This is mainly implemented by the unsupervised decoder.  Bottom-up saliency is needed, since the imagery used to train the top-down detector may only encompass a fraction of the possible targets that may be encountered in real-world scenarios.

There is a major difference between the two branches, aside from from how they are trained.  Never-closed skip connections \cite{SrivastavaRK-coll2015a} from the {\sc VGG}13 networks are added to layers in the unsupervised {\sc SDN}13.  These skip connections serve a dual purpose.  The first is that they promote reliable saliency labeling in the upsampling process.  This occurs because high-resolution features from the encoder are cascaded with the upsampled layer output \cite{PinheiroPO-conf2016a}.  The successive transposed-convolution layers can then learn to assemble target-like region detection and segmentation estimates that are rather accurate despite the lack of manually labeled exemplars.  Edge blurring is also usually avoided.  Using skip connections in this manner additionally shortens the overall training time due to avoiding singularities caused by model non-identifiability \cite{OrhanAE-conf2018a}.  Such connections also routinely preempt a related vanishing gradient problem \cite{HuangG-conf2017a}, which speeds up the learning process.

Batch normalization \cite{IoffeS-conf2015a} is applied to all transposed-convolution layers in both branches.  Its role is to improve the training speed and stability of the network by smoothing the optimization landscape \cite{SanturkarS-coll2018a}.  Leaky-rectified-linear-unit activations are applied after batch normalization to introduce non-linearities into the feature transformation process \cite{NairV-conf2010a}, which, as we noted above, helps to yield good performance.

In some CSAS images, target aspect coverage may be lost, leading to a reduction of contrast and anisotropy and thus target conspicuousness.  The inferred saliency maps may have many mistakes for these regions.  In other CSAS images of the same area, better target aspect coverage may be available.  To counteract a loss of local aspect coverage in any one image, we combine saliency maps from CSAS images of the same area.  Doing so generally improves segmentation quality due to a simulated increase in aspect coverage.  

Combining saliency maps from related areas necessitates uncovering a robust, non-linear image warping and then aggregating the transformed saliency maps.  We do this by first aligning the CSAS imagery via a pre-trained, small-scale version of the {\sc FlowNet} \cite{DosovitskiyA-conf2015a,IlgE-conf2017a} and {\sc LiteFlowNet} architectures \cite{HuiTW-conf2018a}, {\sc LFN}, which is shown in \cref{fig:lfn}.  Such a network extracts pyramidal features that are used to iteratively inform a regularized, dense correspondence map between image pairs.  This correspondence map is progressively refined and smoothed in a way that respects image content.  {\sc LFN}s uncover significantly better flow fields than many alternate correspondence models as a consequence of this functionality (see \hyperref[secC]{Appendix~C}).

Our {\sc LFN} has half the number of upsampling layers as {\sc LiteFlowNet}.  This change reduces the number of parameters and greatly lowers inference time, albeit at the expense of flow accuracy.  We have found, though, that this accuracy loss can be overcome by incorporating additional convolutional layers between each flow decoder, as we have done in \cref{fig:lfn}(a).

Once an image-to-image transformation is found by the {\sc LFN}, previous saliency maps are then warped so that they are aligned.  Structured local predictors \cite{BuloSR-conf2012a} are then applied to facilitate label consensus amongst multiple images.  They take into account local neighborhood appearance, relative position, and the warped segmentation labels to create a unified segmentation mask.  Their results are usually better than ones obtained via purely convolutional means.  The unified mask is then added to the aggregation and refinement sub-network of the {\sc MB-CEDN}, as shown in \cref{fig:mbcedn}(a).

\vspace{0.2cm}\noindent {\small{\sf{\textbf{MB-CEDN Segmentation Post-Processing.}}}} The way that we train the MB-CEDNs causes the decoders to provide disparate interpretations of underwater scenes.  Such interpretations need to be reconciled and converted into a single output response with high spatial consistency.  There may also be regions with spurious labels, especially near the target edges, that need to be corrected.  

To handle these issues, we rely on a deep-parsing network \cite{LiuZ-jour2017a} appended to the decoder branches.  The deep-parsing network efficiently mimics random-field-based structural prediction.  It forces neighboring locations in the combined segmentation map to have similar labels whenever the corresponding features do not deviate much.

Post-processing the target segmentations this way is advantageous.  It conducts conditional-random-field-like inference to approximate the mean-field response in only a single pass.  Multi-pass schemes relying on recurrent layers require significantly more training to achieve robust label consensus.  Far more training samples are also needed.

Once a saliency map is produced, layers could be appended to the {\sc MB-CEDN}s to construct target bounding boxes.  For our experiments, we instead adaptively threshold the saliency maps and use image morphology to remove any small-scale pixel regions that are likely to not contain targets.  The remaining contiguous pixel regions are then isolated and their maximum extents determined, which permits fitting rectangular bounding boxes to them.  This heuristic appears to yield similar performance to using more network layers that conduct bounding-box regression.

\phantomsection\label{sec4.2}
\subsection*{\small{\sf{\textbf{4.2.$\;\;\;$MB-CEDN Network Training}}}}

For the {\sc MB-CEDN}s to prove effective, both decoder branches should transform the encoder-derived contrast features in a way that provides complementary interpretations of the underwater scenes that are then combined into a unified response.   We describe, in what follows, how this functionality can be realized by using a split training strategy with dual cost functions.

\vspace{0.2cm}\noindent {\small{\sf{\textbf{MB-CEDN Supervised Training.}}}} We would like to use human-annotated saliency maps to infer parameters for the encoder branch so that the convolutional layers extract robust, multi-scale contrast features.  Likewise, we would like to use the annotated examples to train the supervised decoder branch so that targets with similar contrast characteristics as the training set are accurately detected.  

\begin{figure*}
\hspace{-0.125cm}\includegraphics[width=6.725in]{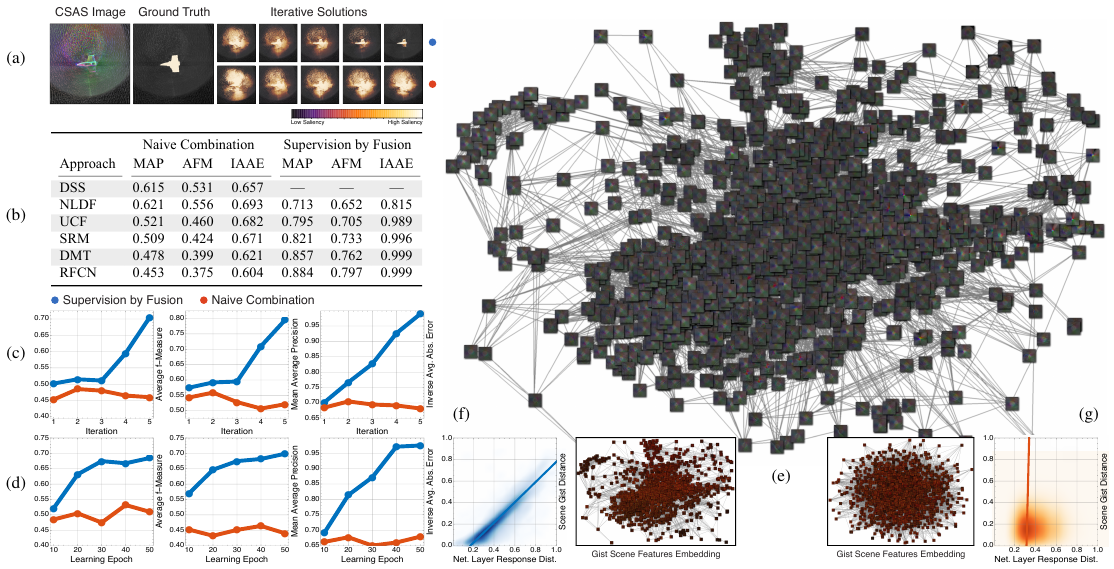}\vspace{-0.1cm}
   \caption[]{\fontdimen2\font=1.55pt\selectfont Illustrations of the disadvantages of directly training the unsupervised branch in a naive fashion.  (a) A CSAS image of a crashed fighter plane and the corresponding segmentation ground truth.  We show two rows of iterative learning results.  The first row (blue) gives the saliency-based segmentation output for supervision-by-fusion learning.  The second row (orange) corresponds to the results after sequentially training on a random weak detector across each iteration.  (b) Statistics for this scene that indicate the improvement with the addition of more weak saliency detectors.  Each row corresponds to the inclusion of that detector with all previous detectors.  We considered six detectors: DSS \cite{HouQ-conf2017a}, NLDF \cite{LouZ-conf2017a}, UCF \cite{ZhangP-conf2017a}, SRM \cite{WangT-conf2017a}, DMT \cite{LiX-jour2016a}, and RFCN \cite{WangL-jour2019a}.  (c) Plots of the average $f$-measure (AFM), mean average precision (MAP), and inverse average absolute error (IAAE), for the scene in (a). Higher values indicate better performance. (d) Statistics plots highlighting the performance across many CSAS images when using only three weak detectors, DSS, NLDF, and UCF.  (e) A $t$-SNE embedding of the encoder features when trained using supervision-by-fusion; here we show $\epsilon$-nearest-neighbor connections in the non-projected space.  We additionally superimpose polar-coordinate representations of gist features \cite{IttiL-jour1998a} for these scenes on the $t$-SNE embedding in (f) to show that the network self-organizes the features by the visual properties of expansion, roughness, openness, and ruggedness.  Supervision-by-fusion hence models well the scene segmentation difficulty.  Both the network features and gist are linearly correlated, as illustrated in (f).  A naive combination strategy is unable to detect targets effectively.  It does not organize scenes based on visual content, as illustrated in (g).  There is effectively no correlation with gist.  We recommend that readers consult the electronic version of the paper to see the full image details.\vspace{-0.4cm}}
   \label{fig:supervisionfusion}
\end{figure*}

Toward this end, we rely on mini-batch-based back-propagation with the following pixel-wise Shannon cross-entropy cost function.  This function measures the overlap between the actual and ground-truth probabilistic saliency scores
\begin{equation*}
\textnormal{max}_\theta\,\sum_{i = 1}^n\sum_{j = 1}^d \Bigg(\frac{\omega_{i,j}}{n}\textnormal{log}(\psi_{i,j}^\theta) \!+\! \frac{(\!1 \!-\! \omega_{i,j})}{n}\textnormal{log}(1 \!-\! \psi_{i,j}^\theta) \Bigg) \!+\! \lambda \|\theta\|_2^2.
\end{equation*}
Here, the mini-batch size is given by $n$, while $d$ is the total number of pixel regions in the output saliency map.  The variable $\lambda$ provides the weighting for an $L_2$-ridge-regularization factor, which helps prevent overfitting.  The term $\omega_{i,j}$, $i \!=\! 1,\ldots,n$ and $j \!=\! 1,\ldots,d$, represents the ground-truth label at a given pixel, while $\psi_{i,j}^\theta$ is the predicted\\ \noindent saliency response, from the supervised branch, at a given pixel, which depends on the network parameters $\theta$. 

An advantage to using cross-entropy is that the magnitude of the network parameter change is proportional to the saliency error.  Large mini-batch errors induce major changes in the filter weights, which helps the encoder to extract progressively more meaningful features during the early learning stages when many mistakes are commonly made.  Moreover, gradients do not vanish in the saturation regions of activation functions.  Adjustments to the filters in the supervised decoder branch can hence be made to continuously refine segmentation boundaries later in the learning process.  Improved learning rates over other costs, like mean absolute error, are typically observed as a consequence of the cross-entropy-loss' behavior \cite{ZhangZ-coll2018a}.

\vspace{0.2cm}\noindent {\small{\sf{\textbf{MB-CEDN Unsupervised Training.}}}} When using the above cost function, the supervised decoder branch will largely be effective at recognizing targets related to those in the ground-truth.  It may, however, fail to detect novel targets, let alone those that are either partly buried or occluded.  We hence want the remaining decoder branch to transform the contrast features in a way that highlights all salient, non-seafloor regions, not necessarily just those that are highly likely to contain targets of interest.

To implement this behavior, we train the second decoder branch in an unsupervised fashion using a fusion-based strategy \cite{ZhangD-conf2017a}.  Given $n$ training images $\Omega_i$, $i \!=\! 1,\ldots,n$, we rely on $m$ salient target detectors \cite{HouQ-conf2017a,LouZ-conf2017a,ZhangP-conf2017a} to gener-\\ \noindent ate pseudo ground-truth $\omega_k$, $k \!=\! 1,\ldots,m$.  We integrate the results from these weak detectors at local and global\\ \noindent scales based on their estimated region-level reliability and difficulty.  Doing so provides an adequate amount of self-supervision for learning good filter weights.  Without estimating such quantities, incorrect segmentation boundaries from one or more weak detectors will propagate through the network and significantly mislead it.  We show this in \cref{fig:supervisionfusion}.

\begin{figure*}
\hspace{-0.125cm}\includegraphics[width=6.8in]{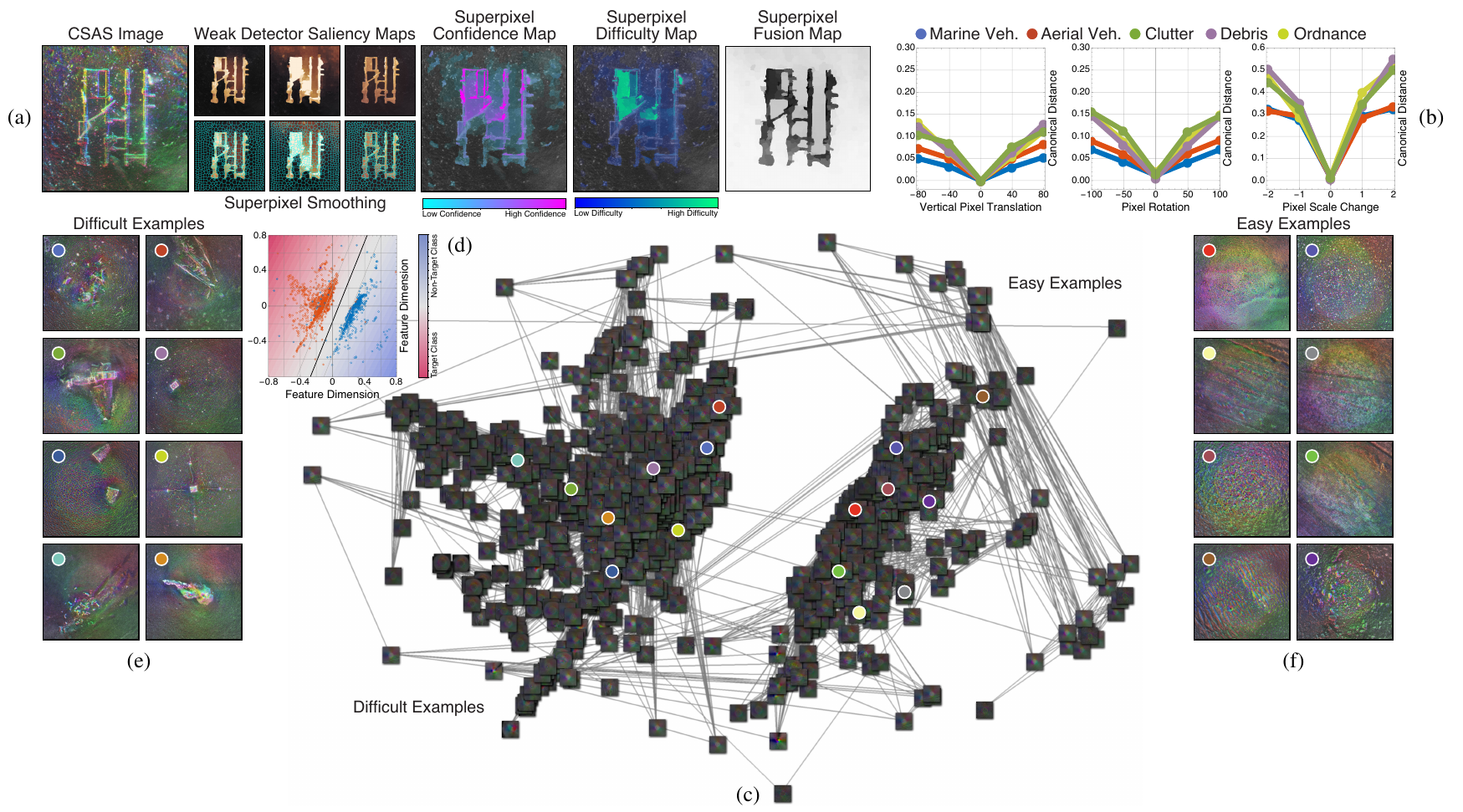}\vspace{-0.1cm}
   \caption[]{\fontdimen2\font=1.55pt\selectfont An overview of the aggregation process and its influence on detection.  (a) For a given CSAS image, candidate segmentation solutions are produced.  Warmer colors indicate a higher target confidence.  Superpixels are then estimated based on local scene content.  The saliency scores within a superpixel region are replaced with the average saliency.  This improves the spatial consistency of the segmentation maps by removing small-scale perturbations.  Local confidence and difficulty estimates are then obtained.  These attributes, along with the saliency scores, are integrated to produce a superpixel-level saliency fusion map.  A similar process is performed at the image level to obtain global fusion maps.  (b) The resulting network contrast features that are learned are insensitive to similitude transformations of the input images.  (c) A $t$-SNE embedding of the local and global difficulty scores for each CSAS image; here, we show the $\epsilon$-nearest neighbor connections in the non-projected space.  The difficulty variables provide a near-perfect separation between scenes that contain targets and those that do not at a global image level.  As shown in (d), a large-margin, linear classifier can separate scenes that contain targets and those that do not.  In (e) and (f), we provide cropped CSAS images of example scenes that, respectively, contain targets and those that do not, along with color-coded indications where they exist in (c).  Scenes with more complexity, and hence are harder to segment, exist toward the left and right fringes of the class distributions.  We recommend that readers consult the electronic version of the paper to see the full image details.\vspace{-0.4cm}}
   \label{fig:csassegmentation}
\end{figure*}

From the weak saliency results, we form a series of local-image fusion maps, $\kappa_i$, $i \!=\! 1,\ldots,n$, that describe\\ \noindent the local reliability of each detector for a single image.  That is, for a decomposition of $\Omega_i$ into $d$ superpixels, the regional intra-image fusion maps, $\kappa_i \!=\! \sum_{k=1}^m a_{i,k} \omega_{i,k}$, dictate how much one detector can be believed over another at labeling a superpixel as either a target or non-target.  Here, $a_{i,k}$ are latent variables representing the reliability of each detector in predicting the correct saliency response for a given superpixel.  Using superpixels reduces the amount of computation.  They also improve spatial consistency of the saliency maps.  The $a_{i,k}$s implicitly depend on a set of difficulty variables $b_{i,j}$, $j \!=\! 1,\ldots,d$, which express how complicated it is for each detector to correctly label a\\ \noindent given superpixel.  Both $a_{i,k}$ and $b_{i,j}$ are inferred using expectation-maximization for a generative model of labels, abilities and difficulties (GLAD) \cite{WhitehillJ-coll2009a}.  An overview of this fusion process, along with corresponding local and global detection results, is provided in \cref{fig:csassegmentation}.

We also combine the weak saliency results into a set of global-image fusion maps, which assess the reliability of the saliency detectors across the global scene content for all images $\Omega_i$, $i \!=\! 1,\ldots,n$, simultaneously.  Similar to the local-image fusion maps, the global ones, $\pi_i \!=\! \sum_{k=1}^m \alpha_k \omega_{i,k}$, rely on a set of image-level reliability variables $\alpha_k$,\\ \noindent $k \!=\! 1,\ldots,m$.  The $\alpha_k$s are inferred along with a set of image-level difficulty variables $\beta_i$ using expectation-maximization within a GLAD model.  Both the $\alpha_k$s and $\beta_i$s depend on a set of labels that reflect the agreement of the weak saliency $\omega_{i,k}$ result with the global average saliency map $\overline{\omega}_i \!=\! \sum_{k'=1}^m \omega_{i,k'}\!/m$ as a function of the normalized $L_1$-distance\\ \noindent between the two maps.

For some underwater scenes, none of the weak detectors may adequately highlight potential targets, leading to poor local and global fusion maps.  To preempt this occurrence, we replace the globally aggregated results that have the lowest prediction reliability with the saliency maps generated by the unsupervised decoder branch.  This yields new weak saliency maps for guiding training.  In the iteration limit, this process typically leads to combinations of detectors with complementary results that detect visually distinct regions well.  \Cref{fig:csassegmentation} shows an example of this.

The local and global fusion maps are employed to estimate the parameters $\theta$ of the unsupervised decoder branch.  This is done via mini-batch-based back-propagation with the following cost function
\begin{equation*}
\textnormal{max}_\theta\,\sum_{i = 1}^n\sum_{j = 1}^d \frac{\gamma_i\beta_{i,j}}{n}\Bigg(\!\Bigg(\kappa_{i,j}\textnormal{log}(\psi_{i,j}^\theta) \!+\! (1 \!-\! \kappa_{i,j})\textnormal{log}(1 \!-\! \psi_{i,j}^\theta) \Bigg) + \Bigg(\pi_{i,j}\textnormal{log}(\psi_{i,j}^\theta) \!+\! (1 \!-\! \pi_{i,j})\textnormal{log}(1 \!-\! \psi_{i,j}^\theta) \Bigg)\!\Bigg) \!+\! \lambda \|\theta\|_2^2.
\end{equation*}
Here, the mini-batch size is given by $n$, while $d$ is the total number of pixel regions in the output saliency map.  The variable $\lambda$ provides the weighting for an $L_2$-ridge-regularization factor, which helps prevent overfitting.  The weight factor $\gamma_i\beta_{i,j}$, $j \!=\! 1,\ldots,d$, represents the elements of the learning confidence map, which is a product of the\\ \noindent normalized superpixel-level confidence map $\gamma_i$ obtained from $b_i$ and the normalized image-level confidence weights $\beta_i$.  The terms $\kappa_{i,j}$ and $\pi_{i,j}$ indicate, respectively, elements of the local-image and global-image fusion maps, while $\psi_{i,j}^\theta$ is the predicted pixel-level saliency response from the unsupervised branch, which depends on $\theta$.

The first scaled Shannon cross-entropy term in this function penalizes the predictions which are inconsistent with the local fusion maps.  The second term penalizes predictions that are not aligned well with the global fusion maps.  The interplay of both terms yields a form of complementary supervision.  It ensures that the saliency maps will exhibit similar characteristics as the fused local and global maps.  

Depending on the choice of weak saliency measures, this branch can highlight targets well as a byproduct of this behavior.  We show this in our experiments.  As with the supervised branch, using a cross-entropy cost helps the unsupervised branch quickly adapt to major errors early during training.  The branch continuously improves from small deviations between the predicted and ground-truth saliency scores at later phases.  This cost function also handles closed-set and open-set noisy labels well \cite{ZhangZ-coll2018a}. The latter naturally manifest under a supervision-by-fusion learning approach. 

Another benefit of this cost function is that it yields a blend of self-paced-learning \cite{KumarMP-coll2010a} and curriculum-learning \cite{BengioY-conf2009a} regimes.  Curriculum learning initially presents the easiest samples followed by increasingly complex ones.  Doing so yields good convergence rates and avoids poor local solutions.  A sample ordering must be pre-specified, though, for it to prove effective.  Self-paced learning infers such an ordering in a data-driven manner.  

There are some differences with our cost function and conventional implementations of both self-paced and curriculum learning.  First, instead of gleaning the self-paced learning weights from a learnt classifier, they are obtained from the fusion process in the form of the reliability variables.  Secondly, rather than pre-defining the sample ordering and fixing it during the entire learning procedure, the ordering is updated across each iteration.  Both changes lead to an adaptive style of training that further improves the convergence rate over standard curriculum learning.  It also aids in uncovering visually distinct regions while ignoring many prevalent seafloor patterns.  This occurs because the training process quickly switches from learning on the easy examples to those with more ambiguity and therefore leads to significant loss reductions.

\phantomsection\label{sec5}
\subsection*{\small{\sf{\textbf{5.$\;\;\;$CSAS Target Analysis Experiments}}}}\addtocounter{section}{1}

We now assess the capability of {\sc MB-CEDN}s for detecting and segmenting targets in real-world CSAS imagery (see section 4.1).  We demonstrate that {\sc MB-CEDN}s can reliably isolate targets, regardless of their appearance and corresponding seafloor type (see \hyperref[sec5.1]{section 5.1}).  We also show that {\sc MB-CEDN}s outperform alternate saliency-based target detection approaches that have been adapted to sonar imagery from natural imagery (see \hyperref[sec5.2]{section 5.2}). 

\setcounter{equation}{0}
\phantomsection\label{sec5.1}
\subsection*{\small{\sf{\textbf{5.1.$\;\;\;$MB-CEDN Performance}}}}

We first illustrate the target analysis behaviors of {\sc MB-CEDN}s before comparing them with alternate approaches.

\vspace{0.2cm}\noindent {\small{\sf{\textbf{Training Protocols.}}}} The {\sc MB-CEDN}s were trained on binary saliency maps using ADAM-based back-propagation gradient descent with mini batches \cite{KingmaDP-conf2015a}.  We relied on the default parameters for ADAM.  We used a mini-batch size of 32 samples to bias against terminating in poor local minima \cite{HardtM-conf2016a}.  

To assess detection quality, we relied on three common metrics, which were mean average precision (MAP) \cite{LiuN-conf2016a}, average $f$-measure (AFM) \cite{AchantaR-conf2009a}, and the average intersection over union (AIOU) \cite{ZhangD-conf2017a}.  For segmentation performance, we used MAP and AFM along with the inverse average absolute error (IAAE). These statistics produce values in the range of zero to one, with higher values indicating better performance.

We pre-trained the {\sc MB-CEDN}s on the 23080-image PASCAL VOC 2012 dataset \cite{EveringhamM-jour2010a,LiY-conf2014a}.  This dataset contains 11530 images of indoor and outdoor scenes with 6929 saliency masks for 20 object classes.  We then fit the {\sc MB-CEDN}s to our 2007-scene CSAS dataset to specialize the networks to imaging-sonar characteristics.  In both cases, we augmented the datasets by performing random translations, rotations, and scalings of the images.  For the sonar imagery, we added multiplicative, normally-distributed random noise to simulate sonar speckle.  Local random haze was sometimes included in the sonar imagery to simulate thermoclines and thus speed-of-sound changes.

As Girshick et al. \cite{GirshickR-conf2014a} have shown, pre-training and then fine-tuning a deep network can significantly improve performance for the target task.  Similar findings have been reported for sonar imagery \cite{HuoG-jour2020a}.  We expand on this point in the online appendix (see \hyperref[secD]{Appendix D}) and show that it is highly beneficial for saliency segmentation.

The performance statistics that we present were averaged across 20 Monte Carlo simulations.  For each simulation, we randomly split and ordered the data into training, testing, and validation sets with ratios of 70$\%$, 15$\%$, and 15$\%$, respectively.  Pre-training on the PASCAL VOC dataset was terminated once the network loss on the validation set monotonically increased for 10 consecutive epochs.  We retained the network weights that yielded the highest validation-set performance out of the 20 random simulations.  We then performed 20 Monte Carlo simulations for the CSAS imagery.  Each simulation defaulted back to the network weights that achieved the best performance for PASCAL VOC.  Training was terminated in the same fashion as for pre-training.

\vspace{0.2cm}\noindent {\small{\sf{\textbf{Experimental Results.}}}} Example detection results for the {\sc MB-CEDN}s are provided in \hyperref[fig:detresults]{figures 5.1}(b)(i)--(ii) and \hyperref[fig:detresults]{5.1}(e)(i)--(ii).  For many scenes, both branches of the network detected the targets well despite the presence of various visual distractors.  The supervised branch was, however, the more adept of the two at isolating previously seen targets and produced tight bounding boxes in such cases.  For target types that were not witnessed during training, the unsupervised branch typically performed better.  It, however, would sometimes erroneously fixated on bottom-type features, such as heavily rippled sand, rock outcroppings, and tilefish burrows, and thus led either to the addition of unnecessary bounding boxes or to boxes with an inflated size.  A combination of the responses from both branches, along with processing multiple CSAS images of the same region simultaneously, resulted in the best detections, as shown in \cref{fig:detresults}(g).

The segmentation results in \hyperref[fig:segresults]{figures 5.2}(b)(i)--(iv) and \hyperref[fig:segresults]{4.2}(e)(i)--(iv) indicate that the {\sc MB-CEDN} branches coherently highlight salient parts of targets well, regardless of their size or complexity.  The unsupervised branch does, however, sometimes erroneously fixate on the seafloors.  This primarily occurs for small-scale targets, like those in \hyperref[fig:segresults]{figures 5.2}(b)(i)--(ii) and \hyperref[fig:segresults]{5.2}(e)(i)--(ii).  In such cases, the amount of acoustic reflectance from the seafloor in the circular full-aspect region is sufficiently greater compared to that in the remainder of the image; this arises from the aggregation of many acoustic pings in this region.  Without either labeled examples to bias against being receptive to such acoustic interactions or more robust underlying saliency approaches used to learn the branch weights, this branch will have difficulties segregating targets from the seafloor.  For larger targets, the issue was less pronounced, since the local reflectances from the targets' facets were much greater than those of the seafloor in the full-aspect region.  Despite this issue, the unsupervised branch did well for all target types, as indicated by \cref{fig:segresults}(g). Performance improved further if the segmentation maps were adaptively thresholded.  The supervised decoder branch achieved even greater target segmentation rates due to better integrating the local and global image features.  As shown in \cref{fig:segresults}(g), the combination of both network branches with label consensus outperformed both.

We also note that the included information on directivity helped with segmentation.  Utilizing reflectivity-only maps reduced average performance by anywhere from one to seven percent for all of the target classes.

\vspace{0.2cm}\noindent {\small{\sf{\textbf{Results Discussions.}}}} Our experimental results indicate that {\sc MB-CEDN}s are capable of accurately detecting and segmenting targets.  In what follows, we describe some of the traits that contributed to its success.

{\sc MB-CEDN}s bear a strong resemblance to saliency-based, FCN-like models for target detection.  {\sc MB-CEDN}s are, however, able to overcome many of the FCNs' flaws, which contributed to their good performance for sonar target analysis.  For instance, unlike FCNs, {\sc MB-CEDN}s do not rely solely on the top-most feature map for inference.  An over-reliance on top-level features may result in poor detection performance on salient regions with weak semantic information \cite{HouQ-conf2017a,LouZ-conf2017a,PhamK-conf2021a}.  {\sc MB-CEDN}s also characterize features at multiple scales, which aids in accurately detecting salient targets of different sizes \cite{YangS-conf2015a}.  This is crucial for sonar target analyses, since target dimensions can vary widely both within and between classes.  FCNs detect targets at only a single scale.  FCN saliency maps are also derived from fixed-size contexts, which often yield blurry target boundaries \cite{LiG-conf2016a}.  {\sc MB-CEDN}s, in comparison, have high spatial consistency near target boundaries, which stems from their ability to learn image-contrast features well.  As hinted at by our experimental results, {\sc MB-CEDN}s can infer both global contrast attributes and edge-preserving, local contrast features.  Incorporating distinct images of the same region also improves performance when one or more of these contexts are inadequate for finding low-contrast targets.

\setcounter{figure}{0}
\begin{figure*}
   \hspace{-0.2325cm}\includegraphics[width=6.65in]{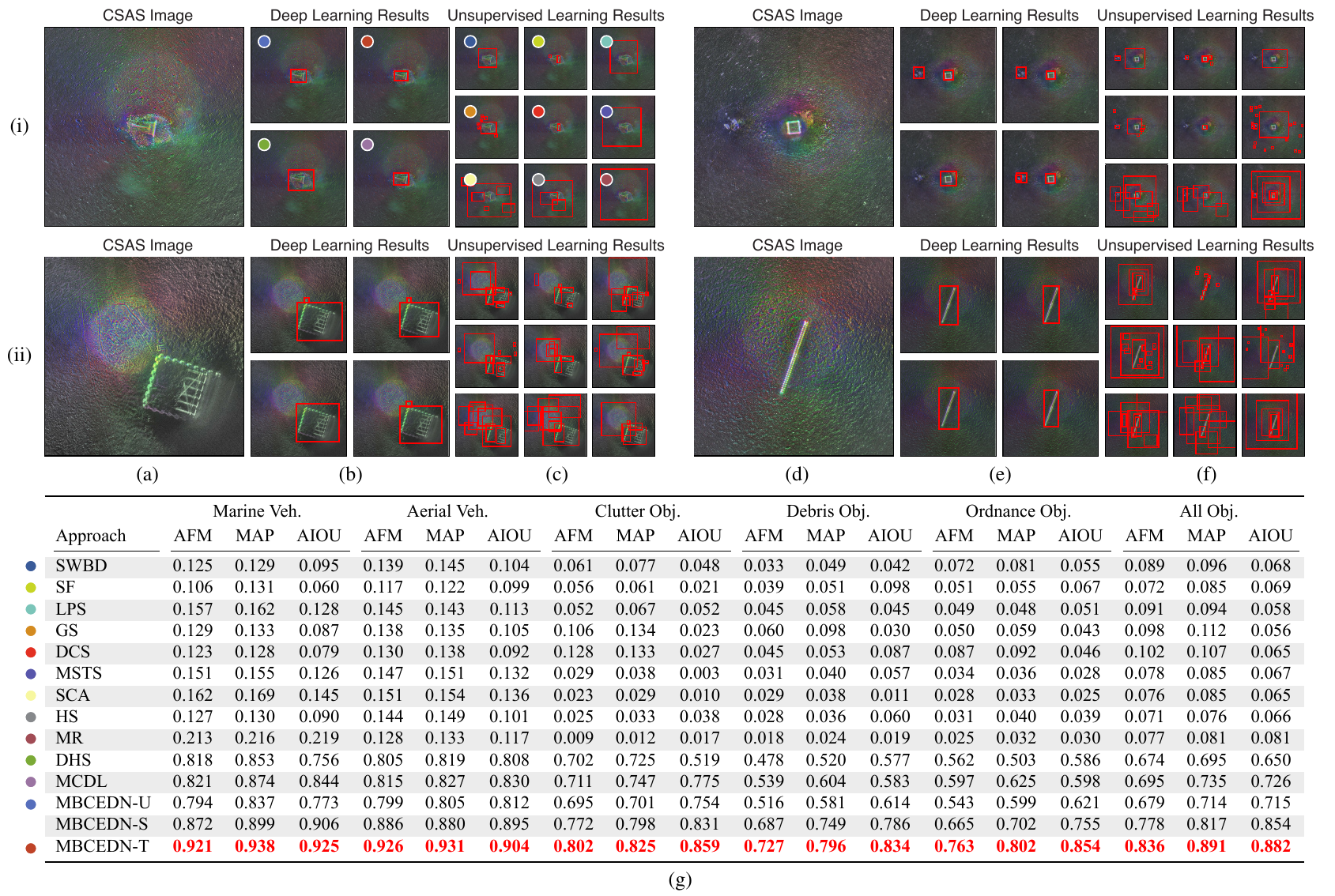}\vspace{-0.15cm}
   \caption[]{\fontdimen2\font=1.55pt\selectfont CSAS target detection results.  In columns (a) and (d), we provide we provide examples of beamformed CSAS seafloor scenes for different target classes.  The top scene in (a) contains a high-holding-power anchor, while the bottom scene in (a) contains an intermodal shipping container.  The top scene in (d) contains a square crate and debris, while the bottom scene in (d) contains a ribbed undersea pipe segment.  These scenes have been cropped near the full-aspect region to emphasize the targets of interest.  In columns (b) and (e), we provide segmentation results for four deep-learning approaches that were applied to the scenes in (a) and (d), respectively.  Going left to right, top to bottom, we show the detection results for the {\sc MB-CEDN} unsupervised branch ({\sc MBCEDN-U}) before reaching the deep-parsing layers, the full {\sc MB-CEDN} ({\sc MBCEDN-T}) executed on multiple images simultaneously, where applicable, {\sc DHS} \cite{LiuN-conf2016a} and {\sc MCDL} \cite{ZhouR-conf2015a}; we do not show the {\sc MB-CEDN} supervised branch ({\sc MBCEDN-S}) branch results, since they do not differ much for these examples.  In columns (c) and (f), we provide detection bounding boxes for nine unsupervised saliency-based target detectors.  Going left to right, top to bottom, they are: {\sc SWBD} \cite{ZhuW-conf2014a}, {\sc SF} \cite{PerazziF-conf2012a}, {\sc LPS} \cite{ZengY-conf2018a}, {\sc GS} \cite{YangC-conf2012a}, {\sc DCS} \cite{YangJ-conf2012a}, {\sc MSTS} \cite{TuWC-conf2016a}, {\sc SCA} \cite{QinY-conf2015a}, {\sc HS} \cite{YanQ-conf2013a}, and {\sc MR} \cite{YangC-conf2013a}.  Note that for each target class, we detect targets in the entire scene, not just the cropped version of it shown in (a) and (d).  In (g), we provide various detection statistics for the different methodologies.  Higher values are better and the best values are denoted using red.  We recommend that readers consult the electronic version of the paper to see the full image details.\vspace{-0.5cm}}
   \label{fig:detresults}
\end{figure*}

\begin{figure}[t!]
   \hspace{-0.275cm}\includegraphics[width=6.66in]{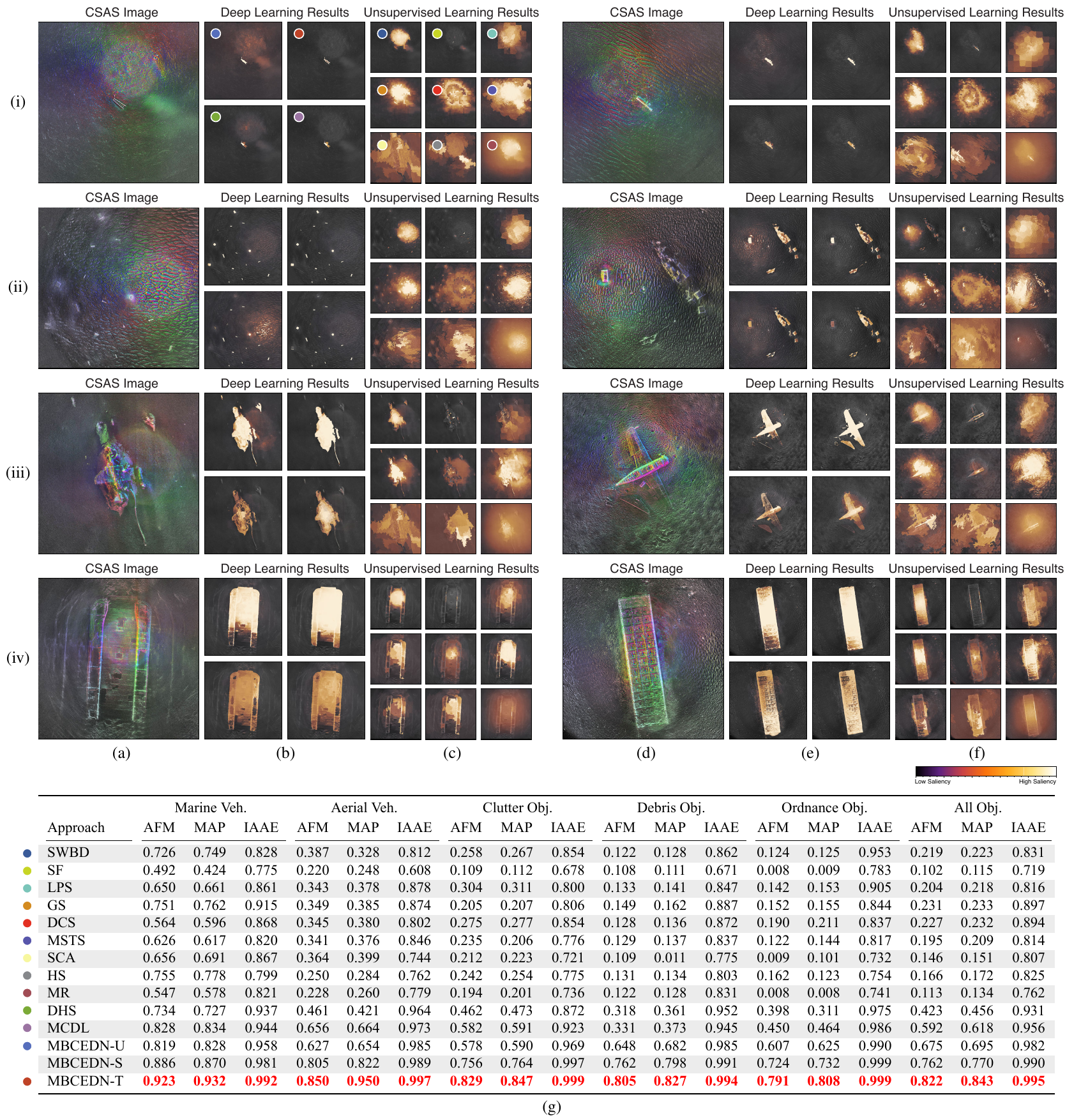}\vspace{-0.15cm}
   \caption[]{\fontdimen2\font=1.55pt\selectfont CSAS target segmentation results.  Each row corresponds to results for a different target class: intact, man-made clutter, man-made debris, aerial vehicles, and marine vehicles.  In columns (a) and (d), we provide we provide examples of beamformed CSAS seafloor scenes for each class.  These scenes have been cropped near the full-aspect region to emphasize the targets of interest.  We have post-processed the images to enhance their characteristics.  In columns (b) and (e), we provide segmentation results for four deep-learning approaches that were applied to the scenes in (a) and (d), respectively.  In columns (c) and (f), we provide detection bounding boxes for nine unsupervised saliency-based target detectors.  For the segmentation images, whiter colors denote a higher confidence that a pixel belongs to a target.  More orange colors denote lower confidence scores.  Note that we segment and detect targets in the entire scene, not just the cropped version of it.  In (g), we provide segmentation statistics for the different methodologies.\vspace{-0.5cm}}
   \label{fig:segresults}
\end{figure}

There are many other reasons why {\sc MB-CEDN}s did well.  In addition to learning robust, multi-scale contrast filters, {\sc MB-CEDN}s adeptly removed nuisances.  They did not fixate on sonar speckle, despite it being prevalent throughout the imagery.  This was partly due to the use of global average pooling operations, which downsample the intermediate feature representations while generally preserving key details and removing random-phase irregularities \cite{WilliamsDP-jour2020a}.  Complex seafloors and marine flora did little to confuse the networks.  Moreover, the encoder convolutional kernels were tuned in a way so that the full-aspect-region reflectances did not greatly impact the contrast features.  The interaction of multiple decoding branches, each trained in a different fashion, also contributed to the overall performance.  When encountering novel target types, especially those whose visual properties diverged from the target training samples, the unsupervised branch often extracted target boundaries better than the supervised branch.  The {\sc MB-CEDN}s weighted the unsupervised-branch responses more heavily in such cases when constructing the final saliency map.  When dealing with previously encountered target types, {\sc MB-CEDN}s emphasized the supervised-branch saliency maps, since they were often superior.  Lastly, the use of a deep-parsing network to post-process the saliency maps enhanced their spatial consistency.  Many deep-learning-based models typically have little to no built-in means of achieving pixel-level target consensus and hence can make many more mistakes.

The way that we constructed and trained our {\sc MB-CEDN}s also aided performance.  For instance, a disadvantage of Shannon cross-entropy loss is it makes pixel-wise independent predictions of the salience probabilities.  The independence property can sometimes cause spatial discontinuities to form in the saliency maps, which manifest as blurry segmentation boundaries.  We largely avoided this issue in our {\sc MB-CEDN}s, which was a byproduct of employing deep-parsing-network layers to force the saliency scores to be spatially consistent for regions with similar contrast details.  It also stems from our use of superpixels in the supervision-by-fusion training of the unsupervised branch.  Super-pixels ensure that the resulting saliency score remains stable within a given cell.  While some segmentation ambiguity is present, it is usually in boundary regions where targets have low-contrast content, which can occur whenever partly buried targets are encountered.  Any saliency measure, without prior knowledge about these types of targets, would have difficulties in correctly labeling such regions.

Additionally, our use of fusion-based supervision permitted the generation of reliable supervisory signals from the aggregation of weak saliency models.  This improved the strength of the pseudo ground-truth and led to better saliency estimates than could have been obtained by any one model.   Other training strategies may not have faired as well.  A naive alternative, for example, would have been to use only a single unsupervised salient target detector to provide initial pseudo-ground-truth and then training the {\sc MB-CEDN} branch by using the saliency prediction results of the current iteration as supervision for the learning iteration.  Only using one unsupervised salient object detector cannot provide adequate supervision, though.  Training deep models under the resulting pseudo-ground-truth maps would inevitably led the network to build trivial feature representations and capture less informative saliency patterns.  Secondly, it lacks a confidence weighting scheme, which plays an important role in guiding the network to gradually combine faithful knowledge from the confident training samples while ignoring the noisy ones.  Due to the different ambiguities of the included contents, different image regions would have different difficulties for obtaining the ground-truth.  Treating all such training samples equally would have introduced non-negligible noise to the learning procedure and led to performance decays.

Pre-training the {\sc MB-CEDN}s on natural imagery also contributed to their success (see \hyperref[secD]{Appendix D}).  Using the PASCAL VOC dataset led to good initial parameters that detected salient targets well.  It also stabilized the upsampling and target-detection filters in both network branches better than the CSAS dataset alone.  It was crucial for regularizing saliency in the deep-parsing-network layers.  Without pre-training, the boundaries specified by the {\sc MB-CEDN}s were not as well-defined as those in the presented examples.  Local contexts were also often ignored, which led to difficulties in identifying the seafloor.  The {\sc MB-CEDN}s additionally had trouble upsampling the saliency maps in a way that respected target edges.

\phantomsection\label{sec5.2}
\subsection*{\small{\sf{\textbf{5.2.$\;\;\;$Comparative Approach Performance}}}}

To provide context for the above results, we compare against deep, supervised architectures and non-deep unsupervised schemes that implement saliency-based target detection.

\vspace{0.2cm}\noindent {\small{\sf{\textbf{Training Protocols.}}}} The first supervised approach we use is deep hierarchical saliency ({\sc DHS}) \cite{LiuN-conf2016a}.  It is one of the earliest and most popular coarse-to-fine deep architectures for detecting and localizing targets in natural imagery.  The second is multi-context deep learning ({\sc MCDL}) \cite{ZhouR-conf2015a}, another detection network that models local and global image features to isolate targets.  Both networks follow a similar encoder-decoder design and extract single-scale contrast features to form an initial segmentation solution that is iteratively refined.  While there are many other available detection architectures, they typically resemble these two.  Based on our prior experiments, are not likely to yield meaningful performance improvements.  

Training of both schemes proceeded as follows.  For {\sc MCDL}, we initialized the underlying {\sc Clarifai} network weights using those from the ImageNet 2013 competition.  The remaining layer weights were randomly chosen.  The network was trained using ADAM-based back-propagation gradient descent with mini batches and data augmentation.  A unified soft-max penalty between the classification result and the ground-truth label was employed as the loss function; this loss was factorized into a product of experts, as described in \cite{ZhouR-conf2015a}.  The hyperparameter values were the same as those for our {\sc MB-CEDN}s.  We also employed the same data splitting strategy as above for the CSAS data. 

For the {\sc DHS} network, we initialized the encoder weights from an ImageNet-trained {\sc VGG}13 network.  The decoder weights were randomly selected.  The entire network was then trained, end-to-end, on the PASCAL VOC 2012 dataset using momentum-based back-propagation gradient descent with mini batches and a cross-entropy loss.  Similar to the authors of \cite{LiuN-conf2016a}, we unfolded the recurrent-convolution layers to avoid the need for training with back-propagation-through-time.  We used a mini-batch size of 12 samples.  Initial learning rates of 0.015 and 0.0015 were chosen for the final layer and remaining layers, respectively.  The learning rate was halved after being presented with 40000 samples.  Momentum and decay factors of 0.9 and 0.0005 were, respectively, utilized.  We then fit the network to our CSAS imagery dataset.  Mini-batch sizes of 5 samples were used.  The initial learning rates for the last and remaining layers were lowered to 0.003 and 0.000003, respectively.  For the remaining parameters, we used the same values as in our {\sc MB-CEDN}s.  We also used the same data splitting strategy as above.

We additionally compare against nine unsupervised saliency-based target extractors.  These include saliency with background detection ({\sc SWBD}) \cite{ZhuW-conf2014a}, saliency filters ({\sc SF}) \cite{PerazziF-conf2012a}, learning-to-promote saliency (LPS) \cite{ZengY-conf2018a}, geodesic saliency ({\sc GS}) \cite{YangC-conf2012a}, dictionary-CRF saliency ({\sc DCS}) \cite{YangJ-conf2012a}, minimal-spanning-tree saliency ({\sc MSTS}) \cite{TuWC-conf2016a}, saliency cellular automata ({\sc SCA}) \cite{QinY-conf2015a}, hierarchical saliency ({\sc HS}) \cite{YanQ-conf2013a}, and manifold ranking ({\sc MR}) \cite{YangC-conf2013a}.  These schemes rely on image appearance and region-connectivity details to distinguish between conspicuous target and background pixels.  Many also integrate image-based priors in an attempt to improve detections in both cluttered scenes and those with low-contrast backgrounds.  

For each of these approaches, we used the same saliency-thresholding and image morphology heuristic as in our {\sc MB-CEDN}s to identify individual pixel regions that were likely to contain targets.  The maximum extents of each region were used to define bounding boxes.  Employing this heuristic allowed for a fair comparison against the {\sc MB-CEDN}s, as performance was driven solely by the underlying saliency-producing processes.

\vspace{0.2cm}\noindent {\small{\sf{\textbf{Experimental Results.}}}} Detection results for {\sc DHS} and {\sc MCDL} are given in \hyperref[fig:detresults]{figures 5.1}(b) and \hyperref[fig:detresults]{5.1}(e) for the four underwater scenes depicted in \hyperref[fig:detresults]{figures 5.1}(a) and \hyperref[fig:detresults]{5.1}(d).  These example scenes show that both {\sc DHS} and {\sc MCDL} can sometimes produce good results.  

Overall, however, both networks lag behind the performance of {\sc MB-CEDN}s, which is apparent from the statistics presented in \cref{fig:detresults}(g).  In some scenes, {\sc DHS} and {\sc MCDL} miss small-scale targets, like spent munitions, that do not have complete aspect coverage and hence are not easily distinguished from the seafloor.  The approaches are sometimes confounded by environmental artifacts, such as rock outcroppings, that appear visually similar to debris and clutter.  {\sc DHS} and {\sc MCDL} also have a propensity to over detect, in the sense that they create multiple, overlapping detection boxes instead of merging them.  An example is shown \cref{fig:detresults}(e)(i), where the debris is unnecessarily fragmented into many components.  Additionally, both networks do not always fit tight bounding boxes around the targets. This can be seen in \hyperref[fig:detresults]{figures 5.1}(b)(i) and \hyperref[fig:detresults]{5.1}(e)(i), where the disturbed seafloor sediment is incorrectly included with the ship anchor and container detection regions, respectively. 

Sample segmentation results for the {\sc DHS} and {\sc MCDL} networks are provided in \hyperref[fig:segresults]{figures 5.2}(b) and \hyperref[fig:segresults]{5.2}(e) for the eight underwater scenes in \hyperref[fig:segresults]{figures 5.2}(a) and \hyperref[fig:segresults]{5.2}(d).  Both networks segmented well when targets had complete aspect coverage.  They highlight the man-made clutter and debris in \hyperref[fig:segresults]{figures 5.2}(b)(i)--(ii) and \hyperref[fig:segresults]{5.2}(e)(i)--(ii), for instance.  Segmentation performance quickly fell when full aspect coverage was lost, though.  Some targets, like the man-made debris and spent munitions in \cref{fig:detresults}(b)(ii) were either missed entirely or only partly captured in these cases.  Larger targets suffered from the same issues.  For the airplane wreckage in \cref{fig:segresults}(e)(iii), parts of the sheared tail fin and wings were not completely isolated, for both {\sc DHS} and {\sc MCDL}, due to both low-contrast acoustic highlights and a lack of total aspect coverage.  When using {\sc DHS}, diagonal indentations in the wings were considered part of the seafloor, which were caused by acoustic shadows.  Acoustic shadows also stymied the segmentation process for the sunken amphibious transport in \cref{fig:segresults}(b)(iv) and the capsized barge in \cref{fig:segresults}(e)(iv); either more training samples or post-processes, such as region-based label consensus, would be needed to correct such mistakes, as we discuss below.  The statistics presented in \cref{fig:segresults}(g) indicate that these types of issues contributed to worse performance compared to our {\sc MB-CEDN}s.

Comparably, all of the unsupervised, saliency-based approaches yielded worse detection results than the deep, supervised networks.  \hyperref[fig:detresults]{Figures 5.1}(c)(i)--(ii) and \hyperref[fig:detresults]{5.1}(f)(i)--(ii) show that they produced many spurious bounding boxes which frequently either missed the targets or predominantly contained just the seafloor.  Such boxes often coincided with ensonified regions where the number of available aspects was changing, which resulted in color-map pattern shifts and hence locally distinct image cues.  For example, for the ship-anchor scene in \cref{fig:detresults}(a)(i) and the water-outflow-structure scene in \cref{fig:detresults}(a)(ii), {\sc SWBD}, {\sc GS}, {\sc DCS}, {\sc HS}, and {\sc MR}, all unnecessarily fixate on the circular, complete-aspect area where the color variation is highly pronounced.  The scenes in \hyperref[fig:detresults]{figures 5.1}(d)(i)--(ii) show that this occurred even if the color-map variation was not overly conspicuous.  Erroneous bounding boxes were also added due the sensitivities of {\sc LPS}, {\sc DCS}, {\sc MSTS}, {\sc SCA}, and {\sc HS} to sonar speckle and acoustic reflectance from small-scale, non-target artifacts; many of these boxes are not displayed in \hyperref[fig:detresults]{figures 5.1}(c)(i)--(ii) and \hyperref[fig:detresults]{5.1}(f)(i)--(ii), since they often encompassed only a few pixels and were removed via post processing.  Moreover, schemes like {\sc SF} and {\sc DCS} typically failed to isolate the entire target, let alone large fractions of it, due to being receptive primarily to edges that arise from high-intensity acoustic returns.  In \hyperref[fig:detresults]{figures 5.1}(c)(ii) and \hyperref[fig:detresults]{5.1}(f)(ii), for instance, the bounding boxes traced the dominant outlines of the water-outflow structure and the ribbed pipe segment, respectively.  All of these issues, and others, contributed to the lower-than-expected detection scores given in figure 5.1(g).

The segmentation plots in \hyperref[fig:segresults]{figures 5.2}(c)(i)--(iv) and \hyperref[fig:segresults]{5.2}(f)(i)--(iv) offer some additional clues as to why the detection rates were poor for these unsupervised schemes.  {\sc SWBD}, {\sc DCS}, {\sc LPS}, {\sc GS}, {\sc MSTS}, and many related approaches can miss small-scale targets, such as discarded steel drums, artillery shells, and exposed piping, even when they are located in the complete-aspect region.  This occurs because the color-map changes are pronounced in this region and hence yield a stronger local-image-cue response than the color-map variations from targets.  Changing the color map did not improve performance.  Different seafloor types, such as rippled sand, also disrupted the segmentation process.  Additionally, the use of compactness and objectness image priors forces the unsupervised schemes to treat small-scale objects as nuisances; they hence are ignored.  Size and location priors, like those used in {\sc MR}, further contribute to the poor segmentation performance of small-scale targets, since they incorrectly assume that targets will be positioned toward the center of the CSAS imagery.  For larger-scale targets, like the sunken amphibious transport in \cref{fig:segresults}(a)(iv) and the capsized barge in \cref{fig:segresults}(d)(iv), many of the unsupervised schemes fared better.  The color-map variation for such targets was often much higher than that for the seafloor, which allowed the local-image cues to more reliably fixate on them.  However, there were so few of these target types, compared to the other classes, which significantly impacted the overall segmentation performance, as shown in \cref{fig:segresults}(g).

\vspace{0.2cm}\noindent {\small{\sf{\textbf{Results Discussions.}}}} The above results indicate that supervised saliency-based target detectors significantly outperformed their unsupervised counterparts.  Moreover, none of the competing methodologies either detected or segmented as well as our {\sc MB-CEDN}s.  There are several reasons why this occurred, which we explain below.

Much like our architecture, {\sc DHS} coarsely detects salient objects from both local and global contexts then proceeds to progressively upsample the estimated saliency map in a hierarchical manner.  A key difference of {\sc DHS}, though, is that saliency-map refinement is iteratively performed using multi-pass recurrent-convolutional layers versus one-pass convolutional layers in {\sc MB-CEDN}s.  The use of recurrent-convolutional layers permits filters in each layer to be modulated by other filters in the same layer, thereby, in theory, enhancing the capture of meaningful details about the context of the targets.  This can result in upsampled saliency maps that preserve well target boundaries and remove superfluous target-detection uncertainties.  In practice, however, {\sc DHS}-like models require a great deal of training to adequately perceive global target properties and avoid the distraction of local interferences.  This happens due to a mixture of gradient vanishing and exploding issues, which require significant tuning to avoid.  The {\sc DHS} networks thus did not generalize as well as the {\sc MB-CEDN}s for segmenting targets.  

{\sc MB-CEDN}s, in comparison, relied on \`{a}-trous transposed convolution layers, not recurrent-convolutional layers, for upsampling.  Such layers require less tuning to converge well.  As well, we used deep-parsing networks.  Deep-parsing networks learn conditional-random-field-like structured predictions without recurrence.  {\sc MB-CEDN}s are easier to train and stabilize quickly.  The target boundaries thus remained consistent during the upsampling process, despite the lack of feedback (see Appendix B).

The {\sc MCDL} network is another hierarchical model for saliency-based target detection.  Akin to our architecture, {\sc MCDL} characterizes saliency in a multi-context fashion.  Global contexts are employed to estimate the saliency of potential targets across the entire image, while local contexts are used for localized prediction near target boundary regions and in areas where there is much textural detail.  In the {\sc MCDL} framework, both contexts are modeled explicitly and integrated whereas the context representations are mixed in the {\sc MB-CEDN}s.  Separating the context extraction should, in theory, yield better-performing detection and segmentation results.  For instance, semantically salient targets should be better localized in low-contrast and cluttered backgrounds when explicitly modeling and optimally consolidating visual cues, background priors, and various high-level target details.  Practically, though, {\sc MCDL}s underperform {\sc MB-CEDN}s due to several factors.  Foremost, {\sc MCDL} networks are distracted by local salient patterns in cluttered backgrounds, due to their sensitivity to sonar speckle.  This stems from the use of average max pooling, versus global average pooling, in the underlying {\sc Clarifai} feature-extraction networks.  {\sc MCDL}s also infer saliency maps in an isolated manner.  Crucial spatial information in the input image and in the intermediate-layer feature maps is ignored.  The inclusion of fully connected layers is to blame for this issue, as they discard spatial coordinates \cite{LongJ-conf2015a} (see Appendix B).  Fully connected layers also add several times more parameters than our MB-CEDNs, which complicates the adaptation from natural imagery to sonar imagery.  Lastly, since all the image patches are treated as independent samples during network training and inference, there is no shared computation among overlapping image segments.  Significant redundancies are present, which do not always aid in detection.

A major contribution to the {\sc MB-CEDN}s' success was the extraction and transformation of multi-scale color, anisotropy, intensity, and orientation features.  This helped the networks to detect targets well, regardless of their size and orientation.  {\sc DHS} and {\sc MCDL} networks, in comparison, only characterize contrast at a single scale.  They have difficulties in accurately isolating small-scale targets, like certain types of debris and ordnance.  They also sometimes have difficulties in making small-scale adjustments to the saliency masks.  Both issues can often be resolved using multi-scale features.

Conventional unsupervised saliency methods for target detection rely on various image cues.  The most widely employed is contrast, which measures the distinctiveness of image regions either locally or globally.  Local contrast methods tend to highlight target boundaries while missing much of the interiors.  Global contrast approaches often uniformly highlight target interiors, thereby leading to better-performing detectors.  They still are unsatisfactory, however, as we saw in our experiments with {\sc SF}, {\sc GS}, {\sc LPS}, {\sc HS}, and other techniques.  It has been previously shown that global contrast fails to preserve important target details.  They often have difficulties detecting small-scale targets and those with complex textures.  This issue is exacerbated when the scene backgrounds are either cluttered or have a similar appearance to the targets.  Many have built-in fixation schemes, which emphasize patterns near the center of the image. Such an assumption is not always appropriate for CSAS imagery, since targets may be present outside the complete-aspect region.  Moreover, these schemes often evaluate contrast using hand-crafted features, like intensity, color, and edge orientation.  The preferred color mapping for CSAS imagery differs great from that of natural imagery, which impedes many of the color-based features.  

Some recent salient detection approaches incorporate prior knowledge into cues in an attempt to improve performance.  Background priors used in methods like {\sc SWBD} encourage regions near image boundaries to be labeled as part of the background.  However, such an assumption can fail when targets have reduced-aspect coverage and appear to blend in with the seafloor.  Compactness priors, used by {\sc SF} and other approaches, favor targets that are compact and perceptually homogeneous elements.  They can fail for large-scale targets, though, like many of the scuttled marine vehicles.  Objectness priors tend to highlight regions which may contain targets of a certain class.  Although such priors can help isolate targets, they are typically implemented using hand-designed features with implicit assumptions.  These assumptions do not always translate well from natural imagery to sonar imagery. They, for instance, fixate on spurious artifacts, like speckle.  Some methods combine cues to utilize their complementary interactions and improve overall performance in certain cases.  These works, however, usually rely on shallow learning models, which make characterizing complicated joint interactions between diverse saliency cues difficult.  Moreover, to preserve conspicuous target details and subtle structures, many methods adopt over-segmentations of images, like superpixels used by {\sc LPS}. Such representations are used either as the basic computational units to predict saliency or as the post-processing methods to smooth saliency maps.  Although these methods can improve saliency detection results for large-scale targets, they may not always be relevant for smaller ones unless the superpixel size can be adapted.

For purely unsupervised methods to be effective for sonar imagery, several changes should be considered.  Foremost, low-level anisotropy features should be extracted alongside multi-scale contrast features.  Such features should derived using properties of observed targets for the chosen color-by-aperture mapping.  They should be regularized by detected edges, so as to respect potential target boundaries.

\phantomsection\label{sec6}
\subsection*{\small{\sf{\textbf{6.$\;\;\;$Conclusions}}}}\addtocounter{section}{1}

We have proposed a novel deep-network architecture for saliency-based detection and segmentation of targets in CSAS imagery.  Our work represents the first automated target analysis effort for such an imaging-sonar modality.  It is one of only a few deep networks for target detection in any imaging-sonar modality.

Our network, the {\sc MB-CEDN}, relies on a convolutional encoder-decoder architecture.  The encoder extracts features that permit segmenting targets.  Dual decoders are used to transform the contrast features into segmentation maps.  The decoders progressively improve the maps by integrating local and global image contexts.  The decoders are trained in a supervised and unsupervised fashion and, respectively, perform well when encountering previously seen and novel target types.

Our experimental results with real-world CSAS imagery showcased the {\sc MB-CEDN}s' capabilities.  They indicated that {\sc MB-CEDN}s outperform conventional deep-network architectures trained in a supervised manner.  This was a byproduct of of using an architecture that derived multi-scale contrast features, which were able to isolate differently sized targets well.  It was also due to using a fully convolutional architecture, enforcing local label consensus, and spatially aggregating features in a way that better preserved saliency details.  Processing multiple CSAS images simultaneously also improved the detection rate when complete-aspect information for a target was lost.

Additionally, our {\sc MB-CEDN}s greatly outperformed unsupervised, saliency-based approaches for target detection.  Such approaches make assumptions that, while applicable to natural imagery, do not extend to sonar imagery.  They assume that targets occupy much of a CSAS image.  This rarely occurs, except for incredibly large targets, like barges, commercial aircraft, and submarines.  Moreover, these unsupervised detectors naively emphasize color-based details, particularly hue transitions.  Non-localized shifts in hue are often indicative of target boundaries in natural imagery.  This is not always true for CSAS imagery, though.  Such changes could be caused by target facet orientation.  They could also stem from differing material types.  A fundamental shift in how these approaches extract and utilize image cues is hence needed to achieve similar behaviors as deep networks for this sonar modality.

Taken together, our findings indicate that alterations to natural-image-based saliency approaches may be needed if they are to perform well for sonar imagery. 

Although our emphasis has been on target analysis for CSAS imagery, the network can, after re-training, also be applicable to SSAS imagery.  In our future work, we will demonstrate that training on the sonar-image sub-apertures from the circular-scan case facilitates transfer learning to the side-scan case.  Little to no re-training may therefore be necessary to achieve good detection and segmentation performance for SSAS, regardless of the target aspect angle.  Moreover, such a transfer-learning process will significantly reduce the amount of human annotation efforts needed to for supervised and semi-supervised training.  Only the CSAS image, not each of the ensuing SSAS-derived images, needs to be annotated.  Any labels can be automatically propagated from the original CSAS image to the sub-apertures via the pixel-level segmentation masks extracted using the {\sc MB-CEDN}s.

CSAS data products offer immense utility for automated analysis of underwater environments.  They possess high shape resolvability, high aspect coverage, and low granular interference levels compared to many SSAS data products.  CSAS data is, however, much more time-consuming to acquire.  Collection times for circular-scan patterns can be greater than those for strip-map surveys by up to a factor of three.  In our future work, we will demonstrate how to gather CSAS-like data at a much faster rate.  We will also develop new techniques for beamforming, autofocusing, and correcting the acoustic signatures.  This will enable dense, large-scale assessments of benthic environments that facilitate a variety of ecological applications.  It will also permit investigators to avoid performing initial SSAS surveys.

\setstretch{0.95}\fontsize{9.75}{10}\selectfont
\putbib
\end{bibunit}

\clearpage\newpage

\setstretch{1.15}\fontsize{10}{10}\selectfont

\phantomsection\label{secA}
\subsection*{\small{\sf{\textbf{Appendix A}}}}
\renewcommand{\thefigure}{A.\arabic{figure}}
\setcounter{figure}{0}

In this appendix, we investigate the qualitative and quantitative effectiveness of color-by-aspect mappings for multi-aspect sonar imagery.  We emphasize that the color scheme used for the CSAS imagery is effective not only for human interpretation, but also automated analysis.

\begin{figure*}
   \hspace{-0.05cm}\includegraphics[width=6.4in]{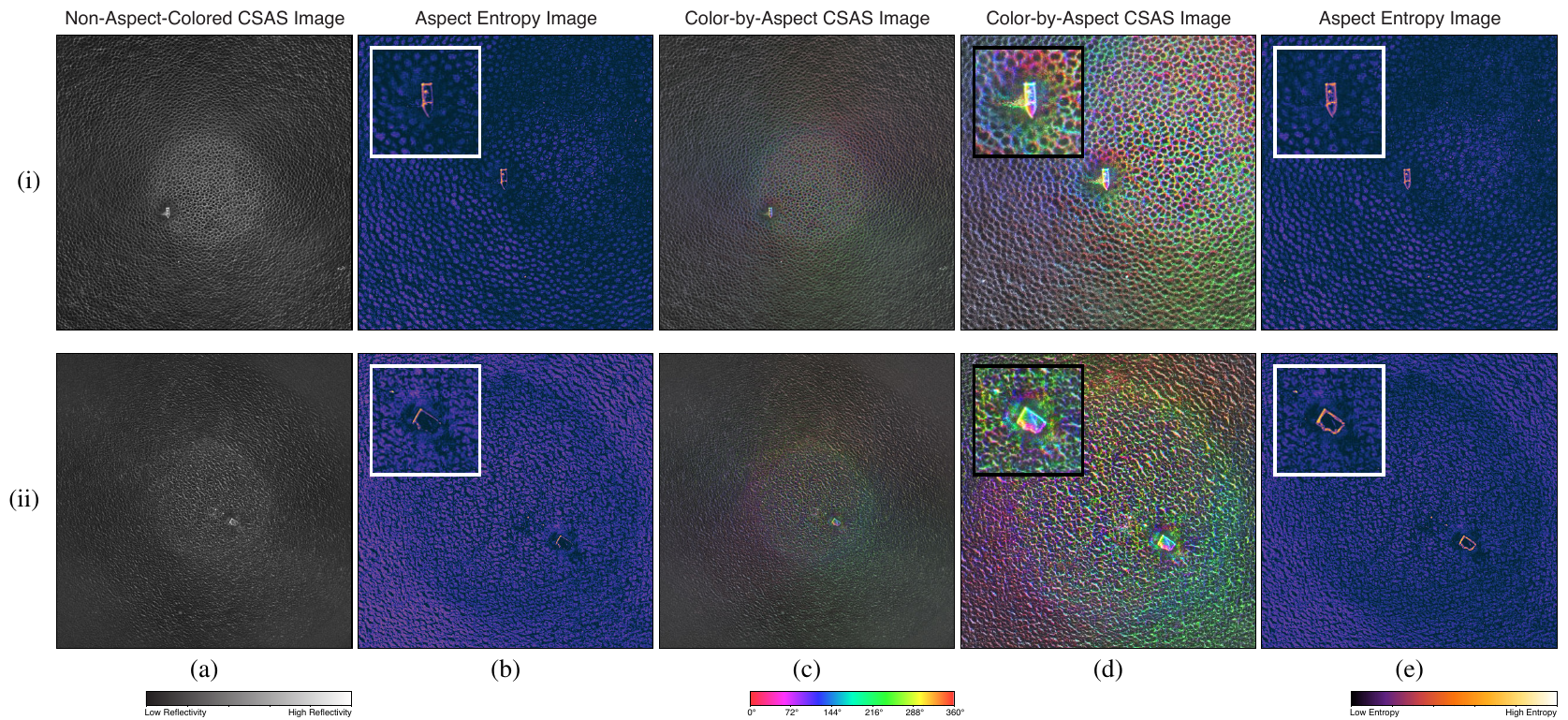}\vspace{-0.05cm}
   \caption[]{\fontdimen2\font=1.55pt\selectfont Examples of two underwater scenes and their corresponding CSAS images.  Row (i) shows a pitted-seafloor environment with an ordnance target, while row (ii) is of a lightly rippled seafloor with debris.  (a) A non-aspect-colored CSAS image.  Here, lightness is indicative of the reflectivity strength.  The color scheme, shown in the bottom-left corner, thus is independent of ensonification angle. (b) A zoomed-in, normalized entropy image of the reflectivity map.  The call-outs in the top-left corner further zooms in on the targets.  (c) An aspect-colored CSAS image, where the angular color scheme is given in the bottom-left corner.    (d) A zoomed in version of (c), with the call-out further enlarging the target.  We applied color, brightness, and contrast transformations to better highlight aspect dependence compared to (c).  (e) A normalized-entropy depiction of (d), showing the target boundaries can be extracted well.  We recommend that readers consult the electronic version of the paper to see the full image details.\vspace{-0.4cm}}
   \label{fig:csas-colormap}
\end{figure*}

To motivate the use of color-by-aspect, we first consider multi-aspect imagery that only contain sub-aperture reflectivity.  Examples are shown in \cref{fig:csas-colormap}(i)(a) and \labelcref{fig:csas-colormap}(ii)(a).  The chosen color scheme is such that dark shades correspond to low aggregate acoustic returns and light shades to high returns.  

While such imagery contains copious textural content that can be used for scene analyses, the lack of directional scattering information can significantly hinder them.  Without an aspect-dependent encoding, target facets can be challenging to distinguish from the seafloor.  For instance, the debris in \cref{fig:csas-colormap}(ii)(a) has a somewhat low total reflectivity, and the corresponding aspect-based entropy image, in figure \cref{fig:csas-colormap}(ii)(b), indicates that there is little change in broadside glint from the surrounding regions.  Both properties suggest that automated approaches would not isolate much of the target well.  Bathymetric details also cannot be reliably assesses for non-aspect-dependent color schemes, which complicates understanding scene topography.  It is difficult, for both \hyperref[fig:csas-colormap]{figures A.1}(i)(a) and \labelcref{fig:csas-colormap}(ii)(a), to discern if there are strong seafloor height changes, for example.

The use of aperture color-coding, as illustrated in \hyperref[fig:csas-colormap]{figures A.1}(i)(c) and \labelcref{fig:csas-colormap}(ii)(c), allows for better interpretation of the scene content.  It highlights bathymetry far more effectively than in \cref{fig:csas-colormap}(i)(a) and \labelcref{fig:csas-colormap}(ii)(a).  It can be discerned, for example, that both environments are mostly flat, with the exception of the indentation around the target.  It is likely that the ordnance recently impacted the seafloor with a modest amount of force, which pushed the sediment and filled in some of the nearby holes.  It appears either that the debris slowly glided toward the seafloor, impacted with its broadside perpendicular to the seafloor and is partly buried, or that sediment has built up around the target over time.  Additionally, the ridges of the pits for the scene in \cref{fig:csas-colormap}(i) are not heavily dominated by directional scattering.  This leads us to believe that they are not rough surfaces and have a graceful gradient.  If they possessed rough edges, like the small ripples in \cref{fig:csas-colormap}(ii), then we would expect to see stronger scattering at lower angles of incidence.

An important insight from \cref{fig:csas-colormap}(i)(c)--(d) and \labelcref{fig:csas-colormap}(ii)(c)--(d) is that targets may be separated from the seafloor not only due to their brightness, but also from their color contrast.  The brightness of a target region is dictated by its acoustic reflectivity.  Color contrast emerges from having facets with strong scattering directions that are different from those of surrounding regions.  Emphasizing anisotropy was a motivating factor for developing the aperture color-coding scheme that we employ.

An additional benefit of using this circular color mapping is that image rotations can be performed by simply circularly shifting the hue wheel.  We take advantage of this trait for data augmentation during network training.

In \cref{fig:csas-colormap}(i)(b) and \labelcref{fig:csas-colormap}(ii)(b), we plot normalized entropy images of the sub-aperture reflectivity maps from \cref{fig:csas-colormap}(i)(a) and \labelcref{fig:csas-colormap}(ii)(a).  Pixels which are highly anisotropic will have entropy values near one.  That is, there is complete uncertainty as to the preferred direction, since sound reflected off of that point for every ensonification angle.  High reflectivity values can also lead to high entropies.  Pixels corresponding to isotopic point scatters will have an entropy near zero.

The entropy maps indicate that portions of the target boundaries can be isolated from the seafloor due to their brightness.  While considering only acoustic reflectivity may be sufficient for defining target bounding boxes, it would not be for contours of complex targets that have varying surface reflectivities.  These variations could be due to material properties or simply the orientation of the target.  Often, it is necessary to consider directionally sensitive features, which help in the latter case.  When considering the aperture-color-coded imagery, the full target boundaries are delineated well in the entropy maps.  Some surface structure is too, like the indentation that separates the munition bourrelet from the body.  This is shown in figures \cref{fig:csas-colormap}(i)(e) and \labelcref{fig:csas-colormap}(ii)(e) and occurs because those facets are either highly reflective, extremely anisotropic, or both.  Such results indicate that, by using a color-by-aperture representation, information is encoded which helps with segmentation.

Note that the aperture color scheme utilized throughout this paper is but one of many possibilities. 

Our decision to associate certain sub-aperture center angles with specific hues was arbitrary.  It has, however, emerged as the dominant color scheme after several human studies.  Investigators can, however, circularly shift the hue color wheel to their preference.  More generally, they can consider any circular color scheme.  

\begin{figure}[t!]
   $\;$\hspace{1.5cm}\includegraphics[width=5.05in]{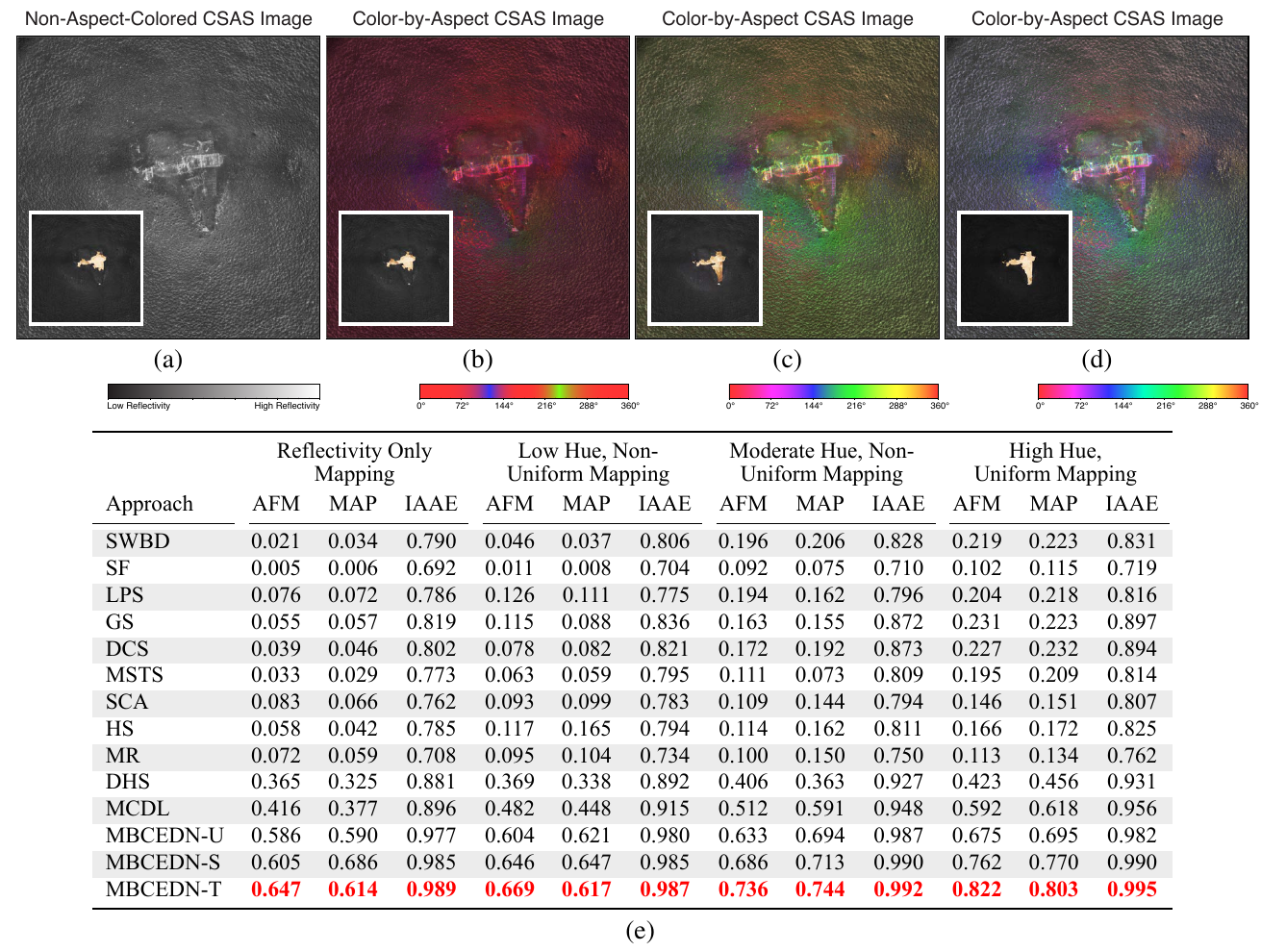}\vspace{-0.1cm}
   \caption[]{\fontdimen2\font=1.55pt\selectfont Illustration of how alternate angular color mappings can reduce segmentation performance.  In (a)--(d), we show CSAS imagery of a crashed airplane.  For (a), only reflectivity is encoded.  For (b), many of the sub-apertures are mapped to red, with only a few to blue.  The mapping becomes more uniform in (c) but it is still compressed compared to the default in (d).  Call-out images in the bottom-left corners highlight that segmentation quality improves for our MB-CEDN as the angular color maps have a wider dynamic range and are increasingly uniform.  In (e), we provide segmentation statistics for the non-angular and angular color schemes considered in (a)--(d).  We retrained the MB-CEDNs for each mapping.  These statistics are averaged across all target classes.  Higher values are better, and the best values are denoted using red.  We recommend that readers consult the electronic version of the paper to see the full image details.\vspace{-0.4cm}}
   \label{fig:csas-altcolormap}
\end{figure}

Compressing the mapping, so that it includes few hues, can significantly hamper the understanding of directional scattering, though.  It can also impact segmentation performance, as we illustrate in \cref{fig:csas-altcolormap}.  In this figure, we consider mappings that progressively encode directional backscattering in a more perceptually distinct manner.  We start from a purely reflective CSAS image in \cref{fig:csas-altcolormap}(a).  We incorporate a small amount of directional sensitivity for the color mapping used in \cref{fig:csas-altcolormap}(b).  In both cases, the lack of anisotropy-based color cues complicates isolating much of the starboard wing.  Only the fuselage and cockpit can be reliably extracted, which is due to their high reflectivity compared to the seafloor.  When the aspect information is emphasized more, as in \cref{fig:csas-altcolormap}(c)--(d), target anisotropy increases.  This permits using color contrast as a discriminative feature for identifying target boundaries.  Solution quality hence improves and is corroborated by the statistics presented in \cref{fig:csas-altcolormap}(e).  These statistics were obtained by using the same training and testing protocols outlined in the experiment section (see \hyperref[sec5]{section 5}).

\clearpage\newpage

\begin{bibunit}
\bstctlcite{IEEEexample:BSTcontrol}

\setstretch{1.15}\fontsize{10}{10}\selectfont

\phantomsection\label{secB}
\subsection*{\small{\sf{\textbf{Appendix B}}}}
\renewcommand{\thefigure}{B.\arabic{figure}}
\setcounter{figure}{0}

\begin{figure*}
   \hspace{-0.125cm}\includegraphics[width=6.85in]{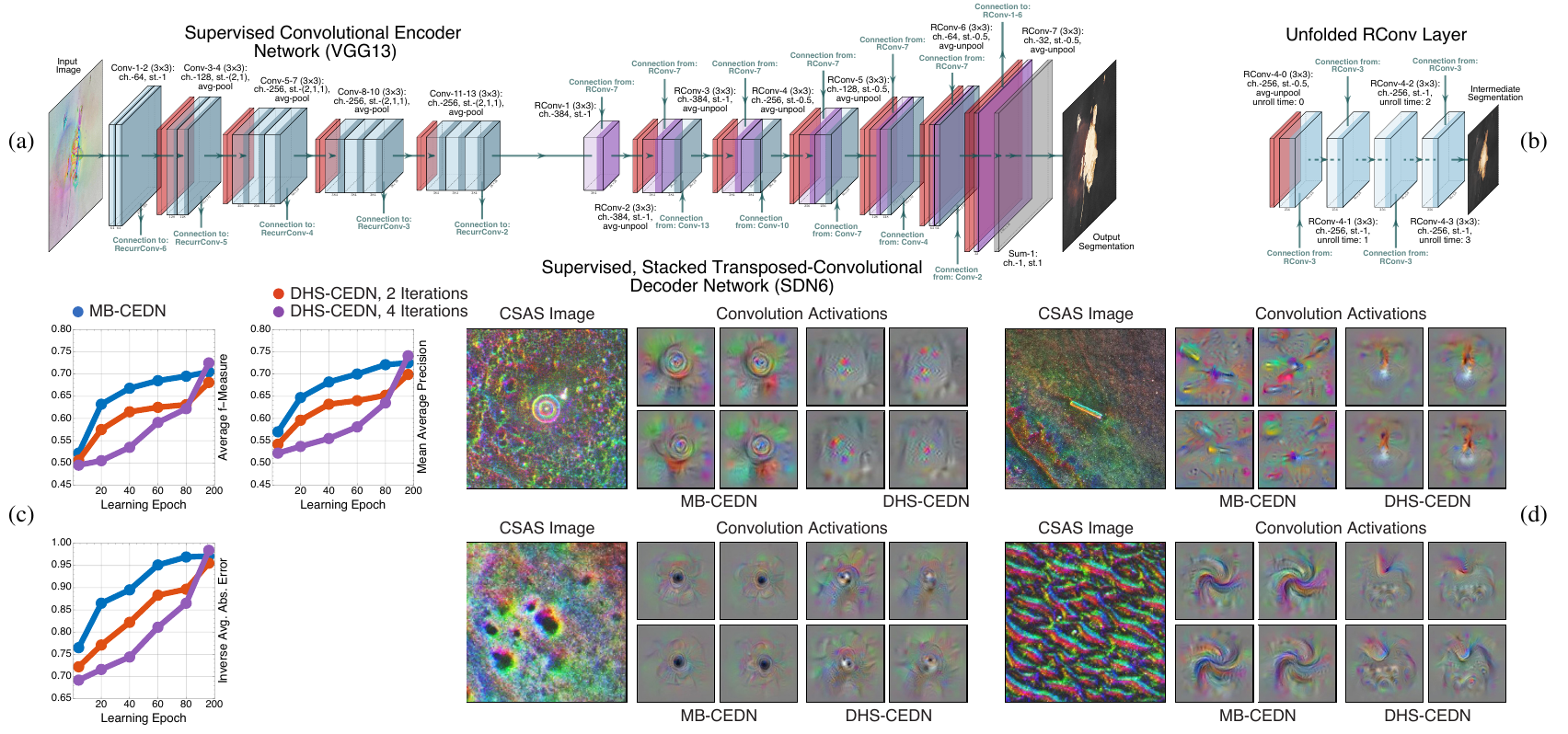}\vspace{-0.15cm}
   \caption[]{\fontdimen2\font=1.55pt\selectfont Illustrations of the disadvantages to using recurrent-convolutional layers in the upsampling process of the saliency map.  In (a), we show a network diagram of a DHS-like version of a single-branch CEDN.  For this diagram, recurrent convolutional layers are denoted using purple blocks.  Here, each of the transposed-convolution layers has been replaced with recurrent-convolution layers that are temporally unrolled for three time steps, as shown in (b).  Note that the last three convolutional layers have shared filter weights.  Dashed lines represent recurrent connections, while solid lines represent feed-forward connections.  Learning statistics plots are provided in (c) for a DHS-CEDN and a supervised-branch only MB-CEDN.  Convolutional activation maps, for a late-stage convolutional filter, are given in (d).  They show the patterns that activate a given filter and hence what the networks look for in a given image.  Here, we consider only the supervised branch of the MB-CEDN.  We recommend that readers consult the electronic version of the paper to see the full image details.\vspace{-0.4cm}}
   \label{fig:dhs-cedn}
\end{figure*}

In this appendix, we further justify some of our architecture decisions, which augments our results discussions.  We quantitatively demonstrate that the differences between the {\sc MB-CEDN}s and existing saliency networks, like {\sc DHS} and {\sc MCDL}, lead to noticeable performance improvements.

When designing the {\sc MB-CEDN}s, we originally considered a recurrent-convolutional architecture, similar to that of DHS, as shown in \cref{fig:dhs-cedn}(a).  Here, we unrolled the recurrent layers for three time steps, as indicated in \cref{fig:dhs-cedn}(b), so that conventional back-propagation could be used.  This is equivalent to training using back-propagation through time on the unrolled network.  We also implemented weight sharing for the unrolled layers, wherein each convolutional layer after the first, in \cref{fig:dhs-cedn}(b), utilized the same weights as the first layer.

Recurrent-convolutional layers \cite{LiangM-conf2015a} have several theoretical appeals over purely convolutional networks.  Foremost, feed-forward networks have no mechanism to modulate responses of earlier layers, except back-propagation-based weight updates.  They hence cannot easily leverage global perceptual context \cite{AlbrightTD-jour2002a}, provided by deeper layers of the network, to guide local context, in earlier layers, and thus correct initial segmentation errors.  The feed-back capabilities of recurrent-convolutional layers provides this functionality, at least over moderate spatio-temporal scales in the case of \cref{fig:dhs-cedn}(b).  Additionally, recurrent-convolutional layers permit increasing the network depth while keeping constant the number of adjustable parameters.  The filter receptive fields are thus enlarged, in an efficient way, thereby facilitating the extraction of complex features for segmentation upsampling and refinement.

In practice, recurrent-convolutional layers can sometimes be tricky to train well, leading to performance issues.  Vanishing and exploding gradients \cite{PascanuR-conf2013a} can occur if the parameter updates are not properly regularized and clipped.  Additionally, using lengthy time iterations, to capture long-range dependencies, is known to complicate gradient back-propagation \cite{LiangM-conf2015a}.  Learning slowdowns are often observed.  Lowering the unrolled-step count in \cref{fig:dhs-cedn}(b) tends to increase the convergence rate at the expense of obtainable short-term performance.

These concerns motivated our use of \`{a}-trous transposed convolution for converting encoder features to saliency maps.  \`{A}-trous convolutions resolve the contradictory requirements between large feature map resolution and large receptive fields.  While we lose the ability for contextual feed-back when employing this type of convolution, we retain much of the parameter efficiency enjoyed by recurrent-convolutional layers.  Additionally, kernel degradation is often prevented \cite{YangM-conf2018a}, which increases training rates.

We compare the performance of these two architectures in \cref{fig:dhs-cedn}(c); the performance metrics were averaged across 20 Monte Carlo trials.  The {\sc DHS}-like network in \cref{fig:dhs-cedn}(a) takes almost twice as long to perform similarly to the supervised branch of the {\sc MB-CEDN} in \cref{fig:mbcedn}(a).  The latter eventually overtakes the former, but this only occurs when changing the early-stopping protocols to allow for periods of little change in the validation-set performance (see \hyperref[sec5]{section 5}).  The hyperparameters must usually be extensively tuned too.  Without either of these adjustments, the {\sc DHS}-like networks converge to poorer contrast filters than our {\sc MB-CEDN}s.  As depicted on the right side of \cref{fig:dhs-cedn}(d), our supervised-branch-only {\sc MB-CEDN}s effectively learn circle detectors, which aid in segmenting axisymmetric targets, like car tires.  Similarly, on the left side of \cref{fig:dhs-cedn}(d), our network has uncovered a pose-invariant notion of linearly shaped targets, such as tubes and casings.  By the same point during learning, the most analogous filters from the DHS-like networks are not adequately stimulated by such geometric patterns, as seen in their activation maps.  They do not detect and isolate such targets well.  Likewise, they have trouble with seafloor patterns.

\begin{figure}[t!]
   \hspace{-0.075cm}\includegraphics[width=6.6in]{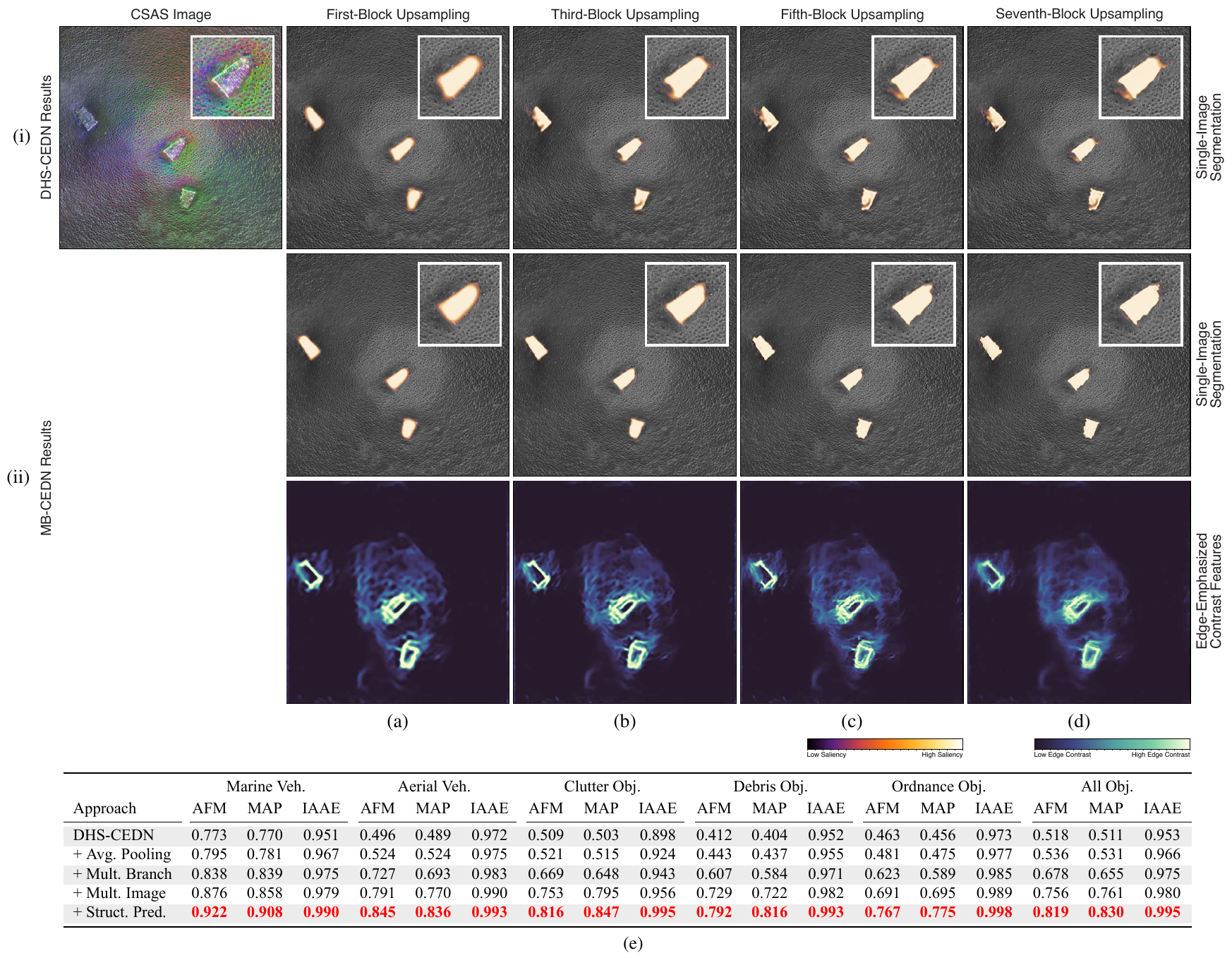}\vspace{-0.105cm}
   \caption[]{\fontdimen2\font=1.55pt\selectfont Segmentation consistency during upsampling for the {\sc DHS}-like network from \cref{fig:dhs-cedn}(a) and the supervised-branch-only {\sc MB-CEDN}, which are shown, respectively, for rows (i) and (ii).  Here, we consider a scene with debris.  In (a)-(d), we provide single-image saliency segmentation maps during the upsampling process.  The maps for each convolutional block have been bilinearly resampled to be the same size as the CSAS image.  In (a)-(d), we also provide aggregate edge-contrast features.  The color scheme is such that light yellow corresponds to a dominant edge while dark blue corresponds to no edge.  The results for the {\sc MB-CEDN} are more consistent during upsampling, despite the lack of recurrent feedback.  They tend to stay within the boundaries specified by the edge-contrast features.  We supply an ablation study in (e).  From these results, we see that the performance of the {\sc DHS-CEDN}s lags behind that of the {\sc MB-CEDN}s.  We recommend that readers consult the electronic version of the paper to see the full image details.\vspace{-0.5cm}}
   \label{fig:dhsupsample}
\end{figure}

Even if good contrast and anisotropy features are learned, segmentation errors can be introduced during the solution generation and refinement processes when using recurrent-convolutional layers.  We show this in \cref{fig:dhsupsample}.  Here, we initialized the encoder filter weights for both the DHS-like network in \cref{fig:dhs-cedn}(c) and the supervised-branch-only {\sc MB-CEDN} in \cref{fig:mbcedn}(a).  We did this using the parameters from best-performing {\sc MB-CEDN} network obtained across 20 Monte Carlo trials; we pre-trained on PASCAL VOC.  We then fixed the encoder weights and trained the decoders using the same protocols as in the experiment section (see \hyperref[sec5]{section 5}).

In \cref{fig:dhsupsample}(a), both the {\sc DHS}-like network and supervised-only {\sc MB-CEDN} have similar initial segmentations after the features are cascaded into the first upsampling block.  They diverge thereafter.  The DHS-like network erroneously fixates on low-contrast regions near the target borders.  While these mistakes, for the target in the call-out of \cref{fig:dhsupsample}(i)(a)--(d), are sometimes corrected as the segmentation solution passes through the remaining upsampling blocks, the lack of a high spatio-temporal span appears to prevent the network from further improving the solutions.  Either additional temporal iterations, additional recurrent-convolutional layers, or alternate recurrent models that can handle long-term dependencies \cite{ZhangL-conf2017a,HansonA-conf2018a} would likely be effective at combating these issues.  By comparison, the large receptive fields implemented by the \`{a}-trous transposed convolution permit the supervised-only {\sc MB-CEDN} to better respect the target boundaries.  Random-field-like regularization helps too.  The solution quality hence quickly improves with each upsampling block, as indicated in \cref{fig:dhsupsample}(ii)(a)--(d).  The intermediate saliency maps also align well with the contrast-feature-derived edges.

These results suggest that there are benefits to considering non-recurrent layers in the {\sc MB-CEDN}s.  The ablation study in \cref{fig:dhsupsample}(e) explores the effects of other architectural changes, each of which positively contributes to segmentation performance.  The largest improvements stem from incorporating multiple, separately-trained branches and implementing structured prediction.  For these statistics, the {\sc DHS-CEDN}s took, on average, twice the number of epochs as the {\sc MB-CEDN}s to converge.  We also had to tweak many of the hyperparameters after adding each new component in the ablation study.

It is important to observe that, unlike {\sc DHS}, the {\sc DHS}-like network in \cref{fig:dhs-cedn}(a) does not possess fully connected layers that constructs an initial saliency segmentation.  Rather, the segmentations are progressively formed across the recurrent-convolutional layers before being softmax-clamped at the output  We, and other authors, have found that fully connected layers impede initial segmentation solutions and thus eschew them \cite{ZhangX-conf2018a,WangT-conf2018a,QinX-conf2019a,ZhaoJ-conf2019a,FuJ-conf2019a,LiY-conf2020a}.  This change explains the discrepancy between the statistics for the base DHS-like model in \cref{fig:dhsupsample}(e) and those for DHS in \cref{fig:segresults}(g).

\begin{figure}[t!]
   \hspace{1.05cm}\includegraphics[width=5.5in]{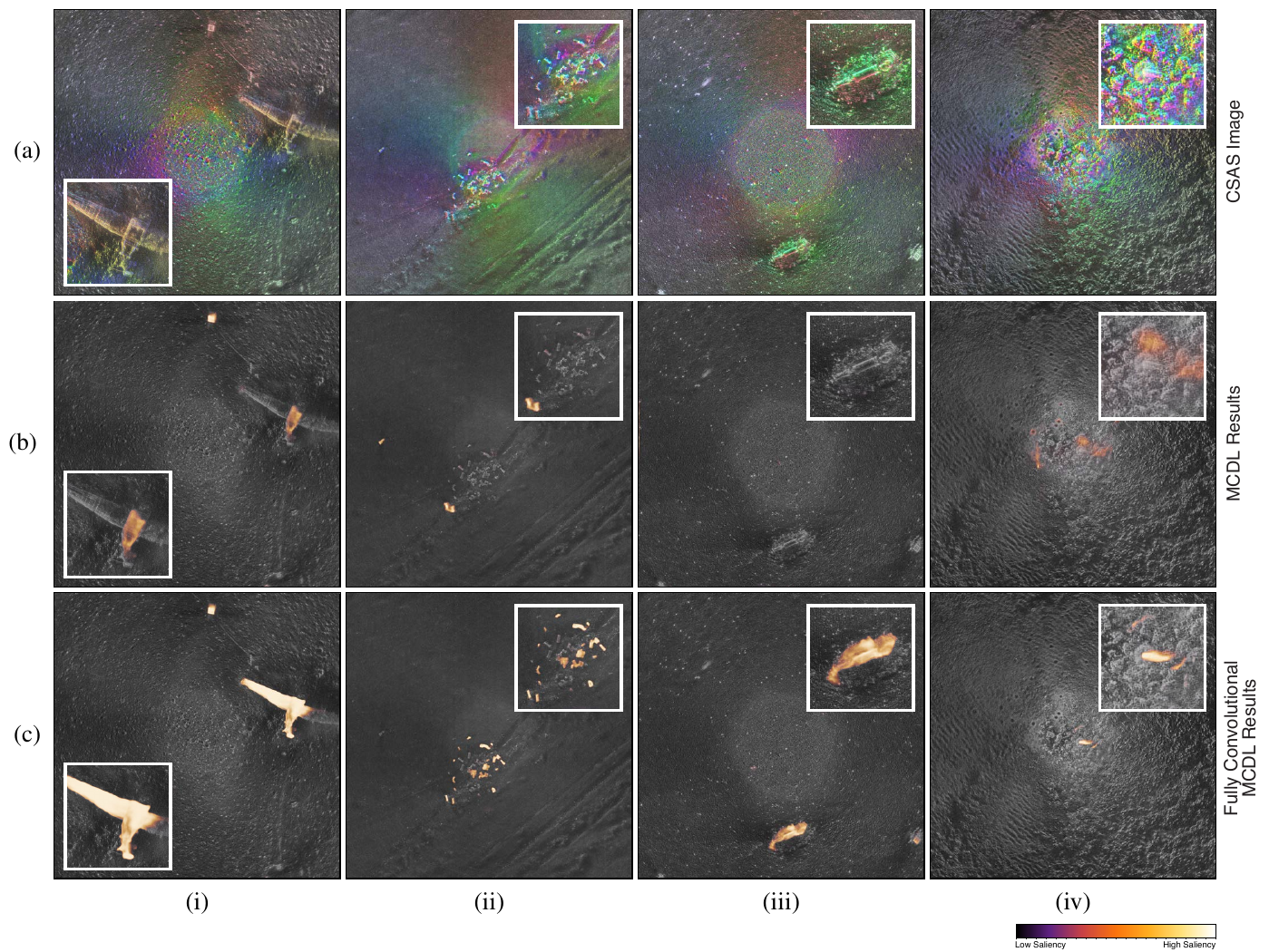}\vspace{-0.05cm}
   \caption[]{\fontdimen2\font=1.55pt\selectfont Issues saliency map specification in {\sc MCDL}.  We consider four scenes in columns (i)-(iv), which are, respectively, of a crashed fighter plane, ordnance, debris, and clutter.  For the CSAS imagery provided in row (a), we visualize the corresponding saliency maps from {\sc MCDL}, in row (b), and an {\sc MCDL}-like architecture without fully connected layers, in row (c).  Here we used the same weights for the encoder backbone and tuned transposed-convolutional blocks attached to the local- and global-context sub-networks.  The results indicate that fully connected layers can ignore spatial information encoded within the convolutional features.  They also miss small-scale targets.  We recommend that readers consult the electronic version of the paper to see the full image details.\vspace{-0.5cm}}
   \label{fig:mcdlerrors}
\end{figure}

Our {\sc MB-CEDN}s were also inspired by {\sc MCDL} architecture.  {\sc MCDL}s contain dual processing pathways that characterize contrast in distinct ways.  One pathway analyzes mostly global context, seeking to characterize contrast in a way that specifies the full extent of target boundaries.  The other pathway considers mostly local context, which yields contrast features that refine the saliency mask to, in theory, better respect small-scale visual details.  Both pathways are trained simultaneously using the same loss.

In our {\sc MB-CEDN}s, we use dual branches to aggregate a mixture of contrast and anisotropy features at local and global scales.  Their aim is to implement top-down saliency, in the supervised branch, and bottom-up saliency, in the unsupervised branch, so that distinct opinions of what is visually conspicuous can be formed and combined.  Top-down saliency should yield low-error saliency maps for targets that are similar to those that have been previously encountered.  Bottom-up saliency should help improve segmentation, both at local and global scales, for novel targets.  Both branches are trained via separate losses so as to promote these behaviors.

Aside from the pathway roles, a major distinction between {\sc MB-CEDN}s and {\sc MCDL}s is how the initial saliency segmentation is formed.  In {\sc MCDL}s, this is done via fully connected layers.  As we noted in the experiment section (see \hyperref[sec5]{section 5}), such layers discard spatial information, often introducing significant saliency errors.  Examples are offered in \cref{fig:mcdlerrors}(b)(i)--(iv) for the images in \cref{fig:mcdlerrors}(a)(i)--(iv), which show that either a majority or the entirety of a target can be entirely missed, regardless of how conspicuous it is.  A significant number of training epochs is needed to overcome this issue.  Replacing the fully connected layers with transposed convolutional layers greatly improved performance, which is qualitatively demonstrated in \cref{fig:mcdlerrors}(c)(i)--(iv).  Early during training, the networks have learned to isolate much of the crashed fighter plane in \cref{fig:mcdlerrors}(a)(i), have found more spent ordnance in \cref{fig:mcdlerrors}(a)(ii), and have delineated the debris and clutter in \hyperref[fig:mcdlerrors]{figures B.3}(a)(iii)--(iv).  Incorporating \`{a}-trous transposed convolution led to further enhancements.

\setstretch{0.95}\fontsize{9.75}{10}\selectfont
\putbib
\end{bibunit}

\clearpage\newpage

\begin{bibunit}
\bstctlcite{IEEEexample:BSTcontrol}

\setstretch{1.15}\fontsize{10}{10}\selectfont

\phantomsection\label{secC}
\subsection*{\small{\sf{\textbf{Appendix C}}}}
\renewcommand{\thefigure}{C.\arabic{figure}}
\setcounter{figure}{0}

\begin{figure}[t!]
   \hspace{-0.075cm}\includegraphics[width=6.4in]{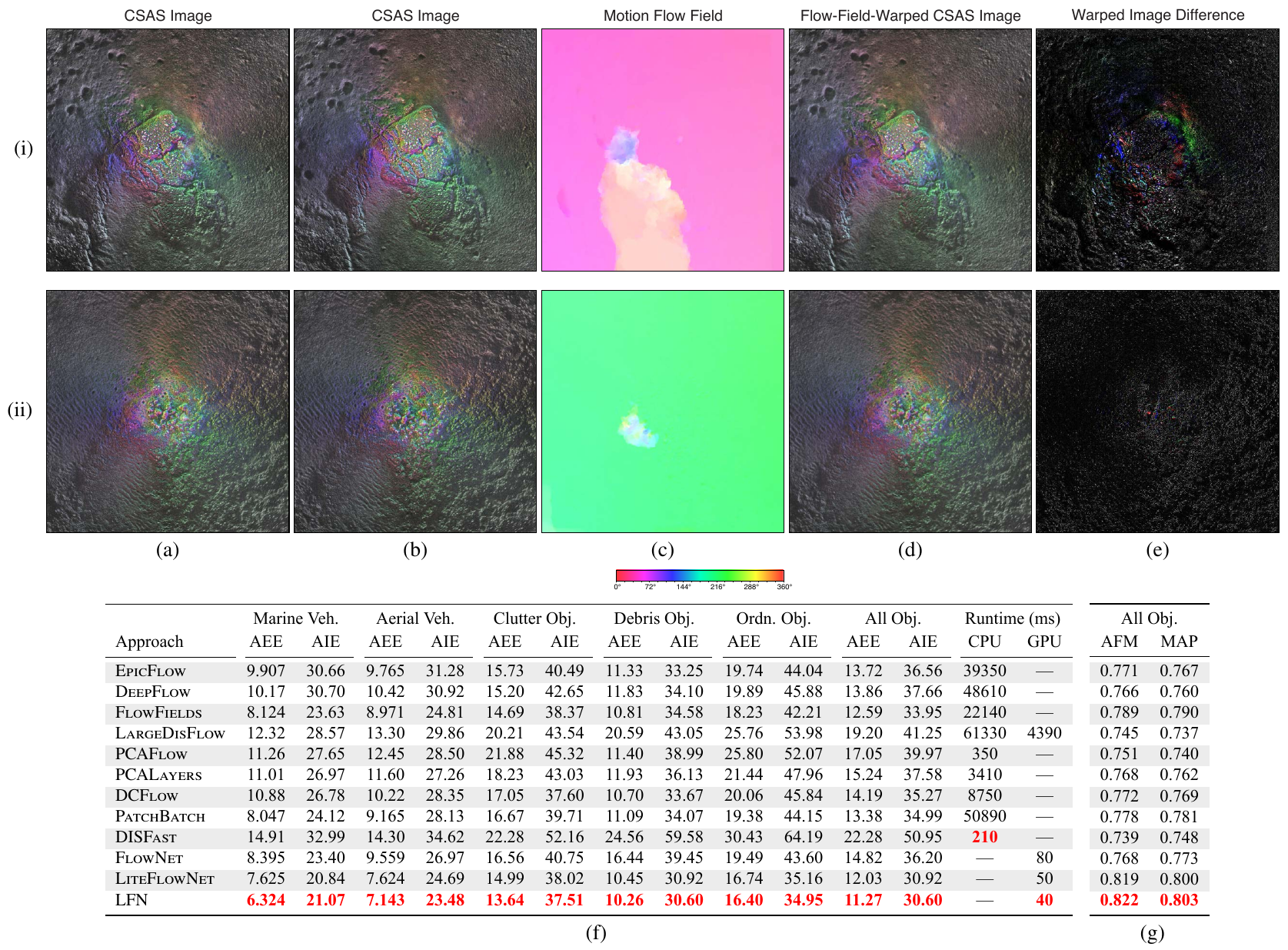}\vspace{-0.105cm}
   \caption[]{\fontdimen2\font=1.55pt\selectfont A depiction of the registration process for multiple CSAS images.  We consider two scenes in rows (i) and (ii), both of which contain rocky outcroppings.  In (a) and (b), we show pairs of images with different center points.  The corresponding {\sc LFN}-inferred flow field between (a) and (b) is given in (c).  Flow vectors are color-coded according to their angle.  The flow-vector magnitude dictates both saturation and lightness.  The images in (d) show how (a) is warped according to the {\sc LFN}-derived flow field.  (d) should resemble (b) if a good flow field is uncovered.  Per-pixel image errors between (b) and (d) are provided in (e).  The errors have been enhanced by at several orders of magnitude for visualization purposes; they are almost imperceptible without this enhancement.  In (f), we provide a table of comparative flow-field statistics.  Here, we compare {\sc LFN} against {\sc EpicFlow} \cite{RevaudJ-conf2015a}, {\sc DeepFlow} \cite{WeinzaepfelP-conf2013a}, {\sc FlowFields} \cite{BailerC-conf2015a}, {\sc LargeDisFlow} \cite{SundaramN-conf2010a}, {\sc PCAFlow} \cite{WulffJ-conf2015a}, {\sc PCALayers} \cite{WulffJ-conf2015a}, {\sc DCFlow} \cite{XuJ-conf2017a}, {\sc PatchBatch} \cite{GadotD-conf2016a}, {\sc DISFast} \cite{KroegerT-conf2016a}, {\sc FlowNet} \cite{DosovitskiyA-conf2015a}, and {\sc LiteFlowNet} \cite{HuiTW-conf2018a}.  We use the average endpoint error (AEE) and average interpolation error (AIE) to quantify performance.  Runtime statistics are also listed.  Lower values are better.  In (g), we give a table of segmentation statistics for the full {MB-CEDN} network when using the comparative optical-flow approaches.  Higher values are better.  These results indicate that {\sc LFN} provides the best performance with the lowest computation time.  This is because it includes regularization and refinement processes that aid in flow-field upsampling.  We recommend that readers consult the electronic version of the paper to see the full image details.\vspace{-0.5cm}}
   \label{fig:opticalflow}
\end{figure}

In this appendix, we establish that the choice of flow-inference mechanism affects CSAS image registration.  

If severe alignment issues are encountered, then there will be little benefit in letting the MB-CEDNs handle multiple images during segmentation.  Fortunately, this does not occur when using {\sc LFN}s within our MB-CEDNs.  As depicted in \hyperref[fig:opticalflow]{figures C.1}(i) and \hyperref[fig:opticalflow]{C.1}(ii), {\sc LFN}s do well for scenes with complex seafloor geometry.  They find suitable, non-linear interpolations that minimize the endpoint error and hence yield low per-pixel differences.  Errors that are observed are often due to changes in either reflectivity magnitudes or the aspect hue, as shown in \hyperref[fig:opticalflow]{figures C.1}(i)(e) and \hyperref[fig:opticalflow]{C.1}(ii)(e).  Neither of these discrepancies can always be mitigated well by image-interpolation methods, since the flow fields only characterize motion, not changes in visual appearance.  However, such errors do not influence saliency-map aggregation within the MB-CEDNs, so reducing them is irrelevant for our purposes.

In \cref{fig:opticalflow}(f), we compare the performance of {\sc LFN}s against several non-deep approaches and deep networks.  The {\sc LFN}s convincingly outperform the alternatives with respect to the average endpoint and interpolation error metrics.  {\sc LFN}s thus return more accurate flow fields that facilitate effective multi-image segmentation, as highlighted in \cref{fig:opticalflow}(g).  {\sc LFN}s have a higher throughput too.  

For these statistics, we trained, where appropriate, the approaches on the MPI Sintel \cite{ButlerDJ-conf2012a} dataset before fitting to the KITTI 2012 \cite{GeigerA-conf2012a} and KITTI 2015 datasets \cite{MenzeM-conf2015a}.  All three datasets are widely used for optical flow estimation.  We used the same training and convergence protocols as in the experiment section (see \hyperref[sec5]{section 5}).  We relied on author-supplied parameter values.  The results were averaged across twenty Monte Carlo trials.  The deep networks were initialized using random weights for each trial.

The improvements we observe stem from multiple mechanisms within the {\sc LFN}s not found in the alternatives.  Foremost, the {\sc LFN}s extract motion-based features.  They employ pyramidal, convolutional encoders which define progressively more coarse features for deeper layers.  Features from these deeper layers permit handling large-scale displacements effectively, even in the presence of drastic illumination changes.  Features from earlier layers are integrated, toward the end of the flow inference pipeline, to address small-scale spatial transformations.  The interplay of both feature types is needed for multi-aspect sonar imagery, since the vehicle center position, along with its roll and pitch, can vary dramatically across circular survey trajectories.  In contrast, approaches like {\sc PCAFlow} \cite{WulffJ-conf2015a} and {\sc PCALayers} \cite{WulffJ-conf2015a}, construct flow fields directly from the images at a single spatial scale.  They can hence be ineffective whenever global, large-scale and local, small-scale transformations are simultaneously encountered and keypoint-matching errors accumulate.  These approaches also cannot always handle global acoustic illumination changes well, since the non-feature-based keypoints are sensitive to visual appearance.

The way that the flow fields are inferred from the features also aids in performance.  For each image pair, the encoder-derived features are used in a series of pixel-by-pixel matching processes at progressively larger scales.  This occurs across three processing blocks within the {\sc LFN}, which match feature descriptors, refine the flow, and regularize it.  For the descriptor-matching block, feature-warping and correlation operations are performed, which provide a point-to-point correspondence cost between images.  A residual flow is then constructed from this cost volume.  The upsampled flow-field estimate and residual flow are summed to account for any changes at the particular current scale that could not be predicted solely through convolution-based upsampling.  This yields relatively accurate flow maps up to that spatial scale.  

\begin{figure}[t!]
   \hspace{0.15cm}\includegraphics[width=6.275in]{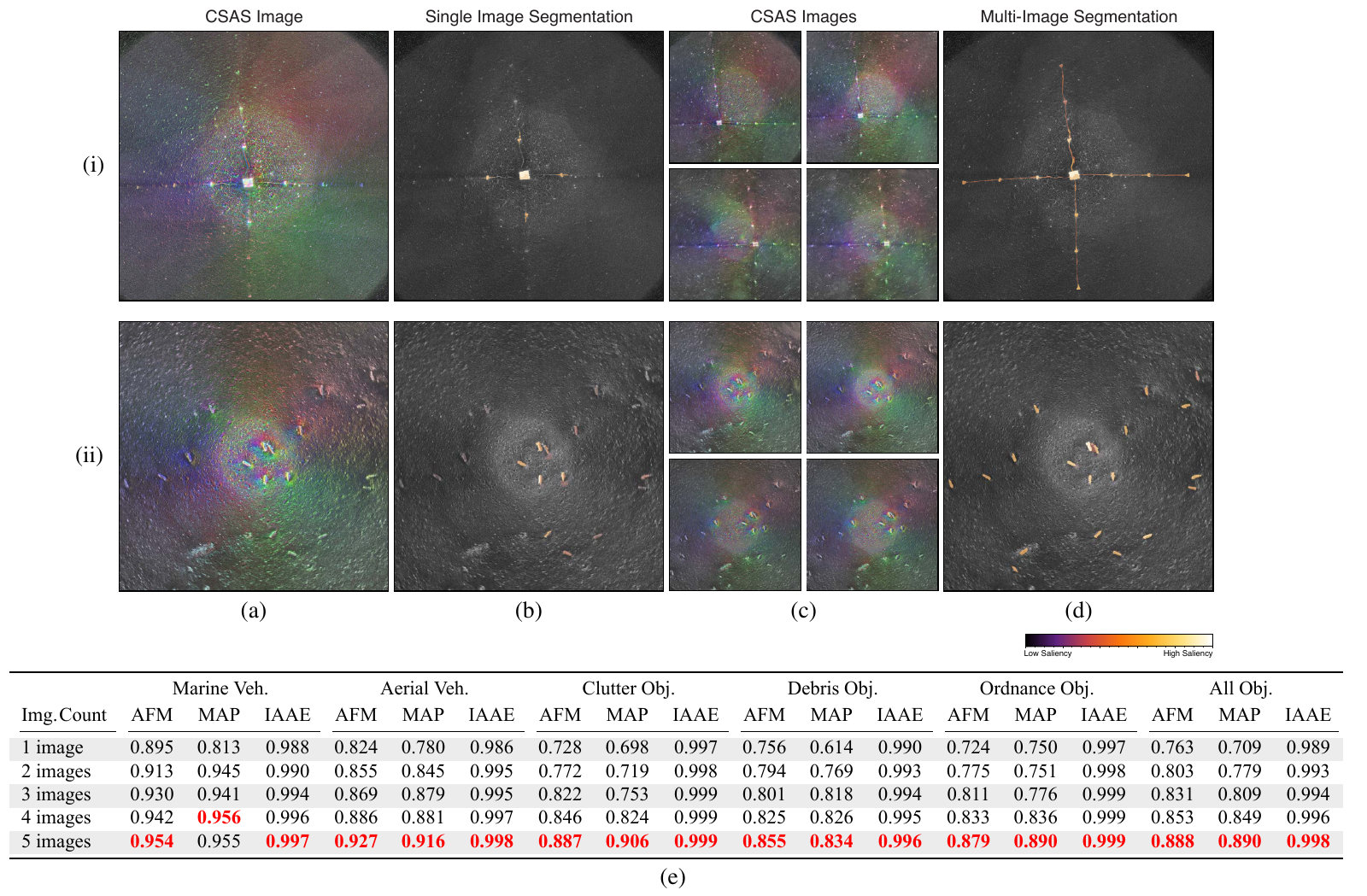}\vspace{-0.125cm}
   \caption[]{\fontdimen2\font=1.55pt\selectfont An overview of the benefits of combining segmentations from multiple images for (i) a localizing marker and (ii) a field of spent ordnance.  In (a), we show a CSAS image, along with its corresponding single-image segmentation in (b).  We utilize up to four additional surveys of the same area, shown in (c), to create an aggregate segmentation (d) that is often better than any single saliency map.  This is due to changes in aperture coverage, which increase the image contrast of certain target regions more effectively in some images.  In (e), we corroborate these claims using several segmentation statistics for the full MB-CEDN network.  Higher values are better.  We recommend that readers consult the electronic version of the paper to see the full image details.\vspace{-0.5cm}}
   \label{fig:multiimage}
\end{figure}

The flow accuracy is further improved by a sub-pixel-refinement network block.  In particular, a secondary residual flow field is computed via minimizing the feature-space distance between one image and an interpolated version of the second image.  Erroneous artifacts are hence de-emphasized when being passed to the next pyramid level.  However, even with sub-pixel refinement, distortions and vague flow boundaries may still be present in the fields.  These issues can disrupt the resulting warped saliency segmentation maps.  {\sc LFN}s remove such details using feature-driven, local convolution within the regularization block.  Such processes adaptively smooth the flow fields.  They act like averaging filter if the flow variation over a given patch has few discontinuities.  It, however, does not over-smooth the flow field across boundaries, thereby preserving well-defined object edges.  Most deep-networks neglect both of these steps \cite{WeinzaepfelP-conf2013a,GadotD-conf2016a}.  In doing so, early-stage errors propagate, disrupting the final flow solution.  The flow fields are also often blurred, especially in the presence of multi-scale motions when using a coarse-to-fine inference strategy \cite{RevaudJ-conf2015a}. 

Due to these beneficial properties, robust flow fields can be uncovered that permit effectively propagating saliency masks from multiple images.  Performance improvements are hence observed for increasing image counts, especially when target-aspect coverage changes greatly.  This is demonstrated in \cref{fig:multiimage}(e).  We obtained the statistics in this figure by training the MB-CEDNs on the 2007-scene dataset used in the body of the paper (see section 3) and evaluating on a 12000-scene CSAS dataset.  The latter dataset is composed of images where up to five circular passes were made per region.  We relied on the same training protocols as the remaining experiments (see \hyperref[sec5]{section 5}).

When using non-{\sc FlowNet} approaches, segmentation performance stagnates when considering more than two images.  The flow fields they return typically have too many issues to effectively transfer segmentation labels, except for CSAS images with slight displacements.  Most of the imagery exhibit moderate to large displacements, though.  Strong ocean currents can shift underwater sensing platforms, causing the center-position to change by several meters.  Likewise, inertial-measurement errors can disrupt vehicle state estimates and lead to the same outcome.  The utility of the non-{\sc FlowNet} approaches that we considered is thus extremely limited for our segmentation application.

\setstretch{0.95}\fontsize{9.75}{10}\selectfont
\putbib
\end{bibunit}

\clearpage\newpage

\begin{bibunit}
\bstctlcite{IEEEexample:BSTcontrol}
\setstretch{1.15}\fontsize{10}{10}\selectfont

\phantomsection\label{secD}
\subsection*{\small{\sf{\textbf{Appendix D}}}}
\renewcommand{\thefigure}{D.\arabic{figure}}
\setcounter{figure}{0}

\begin{figure*}
   $\;$\hspace{1.0cm}\includegraphics[width=5.5in]{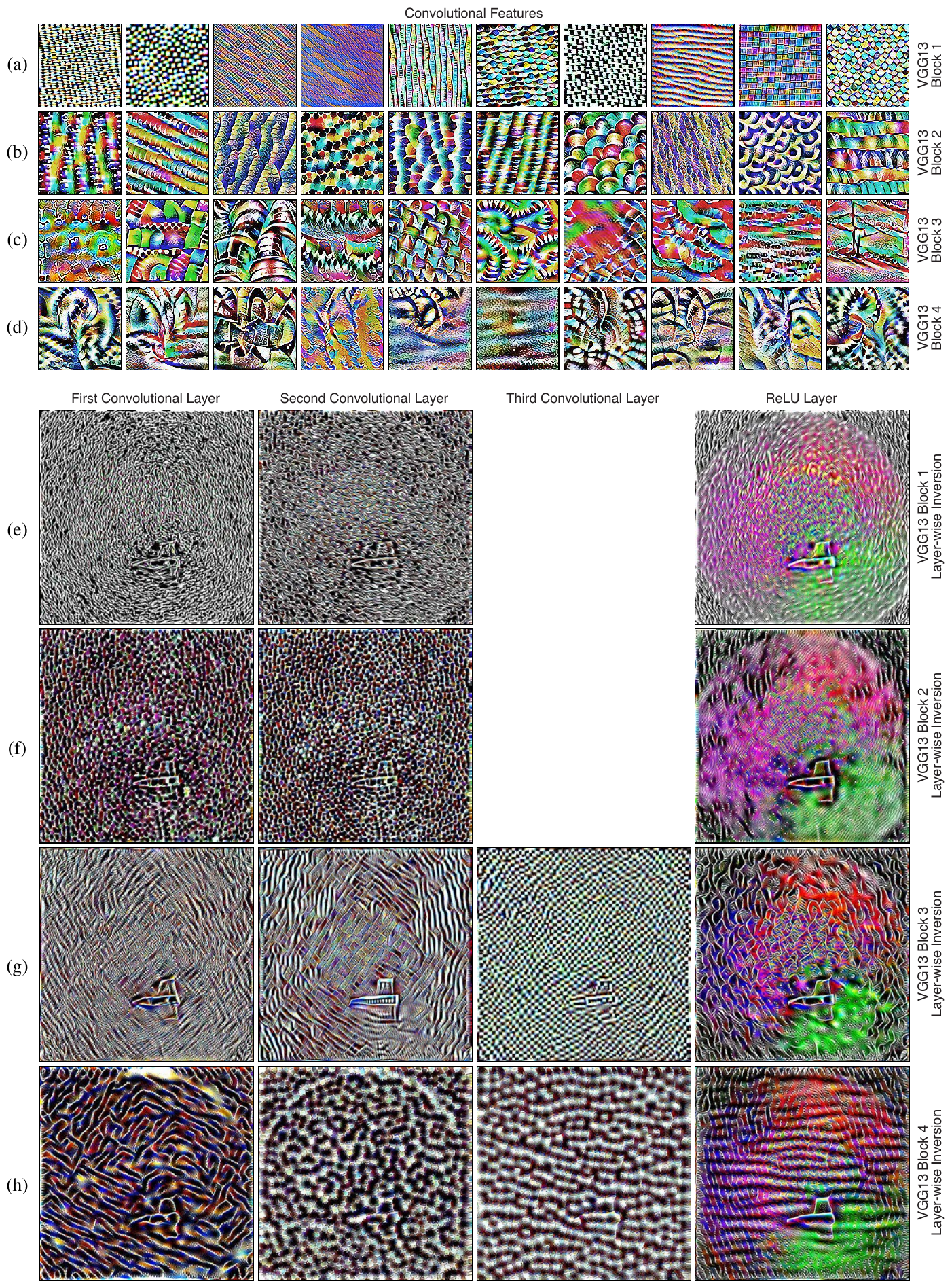}\vspace{-0.1cm}
   \caption[]{\fontdimen2\font=1.55pt\selectfont {\sc MB-CEDN} feature visualizations when only pre-training on the PASCAL-VOC dataset.  In rows (a)-(d), we provide activation mappings from the first four convolutional blocks of the {\sc VGG}13 encoder backbone.  Initial blocks, like (a) focus on directional features, such as lines, and simple semi-periodic textures.  Later blocks, like (b) and (c), are attuned to patterns.  Such components emphasize relevant contrast features for target segmentation.  In rows (e)-(h), we show the aggregate feature inversion \cite{MahendranA-conf2015a} of each layer in the first four convolutional blocks.  Note, that the first two blocks only have two layers each.  We also consider the final rectified-linear-unit ({\sc ReLU}) response for the block.  Here, we consider an image of a crashed fighter plane with sheered wings.  The feature-inverted representations show that deeper blocks of the {\sc MB-CEDN} retain progressively more pronounced directional contrast features used to define target boundaries.  We recommend that readers consult the electronic version of the paper to see the full image details.\vspace{-0.4cm}}
   \label{fig:pretraining-features}
\end{figure*}

In this appendix, we discuss the contrast features that emerge when pre-training the {\sc MB-CEDN}s on the PASCAL-VOC dataset.  We also describe how these natural-image-derived features aid in segmenting sonar imagery.

The first layer of {\sc VGG}13 mainly constructs two types of low-level features, oriented color-contrast detectors and Gabor-like edge detectors.  Both yield a visual reconstruction basis.  Color-contrast detectors react to a particular color on one side of the receptive field and a disparate, usually complementary, color on the other side.  They emphasize localized color transitions for various orientation and shift configurations, which form the foundation of texture in both natural and sonar imagery.  They also highlight target anisotropy when applied to multi-aspect sonar imagery.  Gabor filters mainly respond to edges at different angles.  Other instances of these filters are negative-reciprocals, which fill holes.  These two classes of Gabor filters identify possible object boundaries in a coarse way.  They focus on possible target boundaries, based on the acoustic reflectances, for the imaging-sonar case.  They also facilitate the formation of complex filters in later layers.

The second and third convolutional layers extract a more diverse set of features.  There are invariant color-contrast filters.  There are low-frequency edge detectors, which accent hazy, poorly defined edges of objects.  Such filters are immensely useful for sonar imagery, as errors during auto-focusing can blur target boundaries.  Target facets not acoustically illuminated from multiple directions can also appear blurred, especially if they are near the image fringes.  Color and multi-color filters are also present.  These track local brightness and hue.  Others respond to mixtures of colors, albeit without any obvious spatial preferences.  The remaining features tend to be derived from combinations of first-stage Gabor filters.  The combined-Gabors filters are fairly invariant to exact position and respond to color contrasts that align with the edges.  They appear to be tuned to reveal background distractors composed of local, semi-repeating patterns.  Examples include foliage in natural imagery and rippled sediment on the seafloor in sonar imagery.  The combined-Gabor filters additionally isolate complicated, non-locally-linear target boundaries in a lightness-invariant manner, which helps segment targets in different configurations with variable scattering intensities.

Simple shape predecessors emerge in the third and fourth convolutional layers.  Line, combed-line, shifted-line, curve, shifted-curve, circle, angle, corner, and divergence detectors can be found.  All of these filters react to complete and incomplete versions of their corresponding shape primitives.  Their role is to specify features that reveal object, and hence target, boundaries more reliably than low-level Gabors.  These filters also are the foundation of patterns.  As shown in \cref{fig:pretraining-features}(a), they yield hatched, directional detectors in the activation space.  Rhythmic, wavy detectors are also observed, as are black-white detectors.  These types of activations, along with the other observed patterns, predominantly focus on local changes in frequency, color alterations, and actual edges.
 
\begin{figure*}
   $\;$\hspace{1.0cm}\includegraphics[width=5.5in]{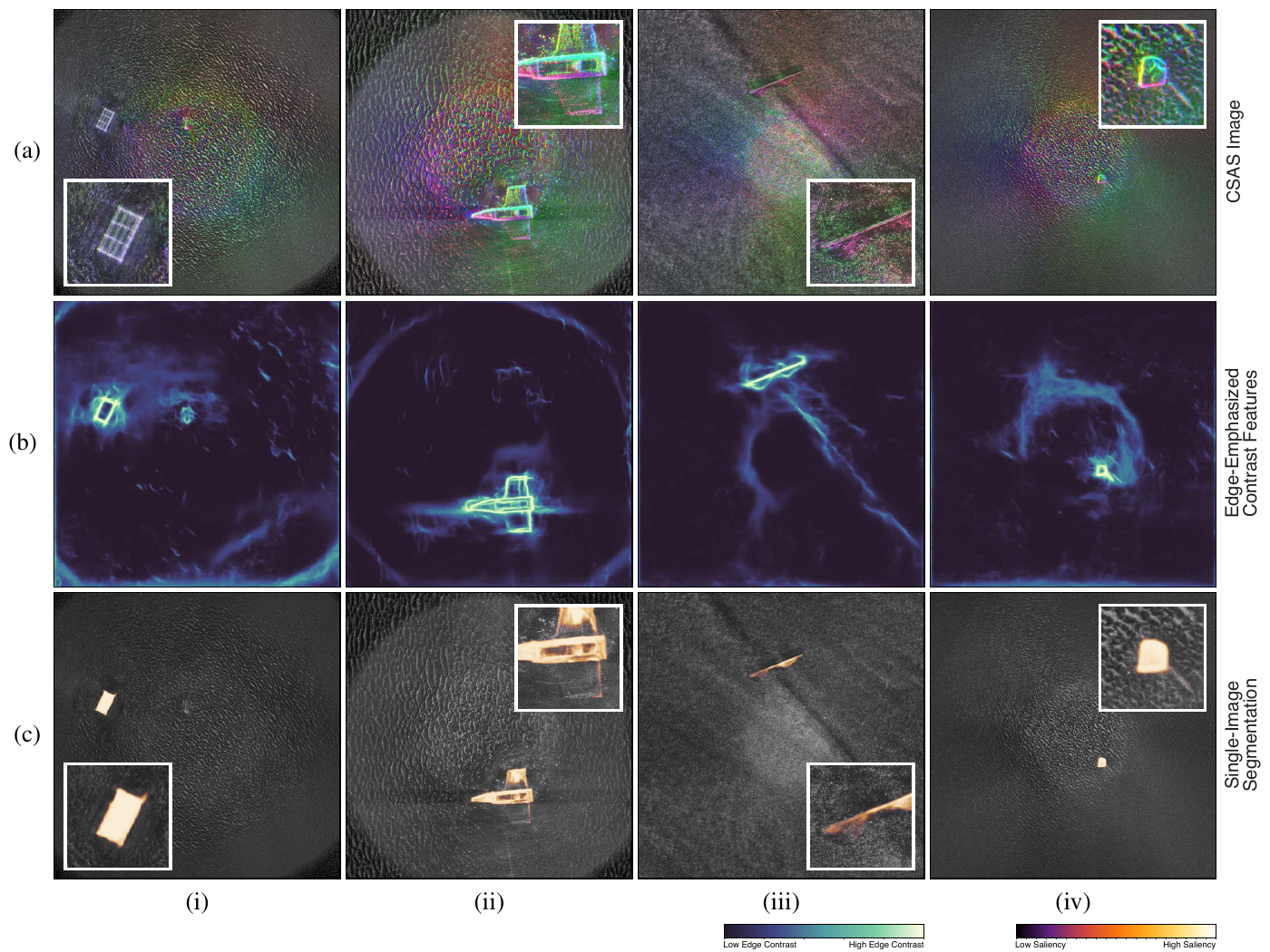}\vspace{-0.1cm}
   \caption[]{\fontdimen2\font=1.55pt\selectfont Pre-trained {\sc MB-CEDN} segmentation results for four scenes, shown in columns (i)-(iv).  In row (a), we provide CSAS imagery for each scene.  Aggregate edge-contrast features are provided in row (b), which highlight the edges that are emphasized by the entire {\sc VGG}13 backbone network.  Segmentation maps are given in row (c).  The results indicate that the PASCAL-VOC-trained {\sc MB-CEDN}s can segment CSAS imagery well.  We recommend that readers consult the electronic version of the paper to see the full image details.\vspace{-0.4cm}}
   \label{fig:pretraining}
\end{figure*}

\hyperref[fig:pretraining-features]{Figures D.1}(e) and \hyperref[fig:pretraining-features]{D.1}(f) illustrate the cumulative effect of these filters for the sonar image presented in\\ \noindent \cref{fig:pretraining}(a)(ii).  Here, we invert the features formed by all of the filters for a given operation in a block of layers \cite{MahendranA-conf2015a}.  At the first-block rectified-linear-unit, the pre-trained {\sc VGG}13 network accentuates the sand-ripple crests, which is a byproduct of the curve, line, contrast, and pattern filters.  The valleys appear to be excluded entirely due to the effects of the black-white filters.  As well, the network isolates, from the seafloor, the plane cockpit, fuselage, and parts of the wings that remain.  Many distractor elements are still present.  These, however, are partly inhibited at the rectified-linear-unit for the second block.  Much of the airplane is distinguished better as a consequence of brightness-gradient filters that fire due to specularities.  The starboard wing is not, though.  The acoustic pings are highly scattered for this target facet, leading to a faint response that is incredibly difficult for many of the line filters to spot.  Such filters would need to be adapted to sonar imagery to account for these kinds of visual characteristics.

Both simple and complex features are observed in the fifth through seventh layers.  There are center-surround detectors, which look for a particular color in the center of the receptive field and another color at the edges.  More elaborate versions, that emphasize center patterns, are also found.  The latter type of filters are sometimes combined with brightness-gradient and color-contrast ones to yield high-low frequency units that highlight texture transitions.  High-low-frequency changes are often an additional cue for object boundaries and help in cases where the objects are juxtaposed against a high-frequency background pattern.  All of the above filters are aggregated in the sixth and seventh layers, leading to pattern and texture activations, which we present in \cref{fig:pretraining-features}(b).  Based on activation-grid images, we conclude that the texture filters recognize object materials in natural imagery.  Pertinent instances include fur, gravel, rocks, tree bark, metal grates, and fabric patterns, many of which correspond to parts of non-salient objects.  For sonar imagery, the same filters fire in the presence of rocky and hard-coral bottom types, pitted sand, along with fields of loose rocks, signifying that they are also indicating the presence of non-salient regions.

Filters for the remaining encoder layers have activations corresponding to intricate, often non-repeating textures.  Some, like in \cref{fig:pretraining-features}(c), appear to be instances of object parts from distinct viewpoints.  Others, such as those in \cref{fig:pretraining-features}(d), are not easy to interpret.  They are amalgams of part and texture detectors and likely serve to characterize object-background interactions for challenging scenes.

As demonstrated in \hyperref[fig:pretraining-features]{figures D.1}(g) and \hyperref[fig:pretraining-features]{D.1}(h), filters from the fifth through seventh convolutional layers remove many distractors that remain after the fourth layer.  Only the most dominant edges for \cref{fig:pretraining}(a)(ii) are retained, which aids in specifying initial saliency maps for conspicuous targets with high reflectivity contrast.  Targets with low anisotropy and low reflectivity contrast often have a weak presence in the initial map.  They require the integration of features from earlier layers to overcome the removal of low-contrast cues, at least in the pre-trained networks.

Despite being solely trained on natural imagery, the {\sc VGG}13 backbone of our {\sc MB-CEDN} does surprisingly well for single- and multi-image segmentation of sonar imagery.  As shown in \hyperref[fig:pretraining]{figures D.2}(i) and \hyperref[fig:pretraining]{D.2}(iii) it does well for targets outside of the complete-aperture region.  The result in \hyperref[fig:pretraining]{figure D.2}(iii) is rather promising, given the challenging target shape.  That in \hyperref[fig:pretraining]{figure D.2}(iv) is too, due to the target size.  Both solutions are better than those returned by the sonar-adapted {\sc DHS} and {\sc MCDL} networks.  Complicated targets, like those \hyperref[fig:pretraining]{figure D.2}(ii), present some challenges, though.  The {\sc MB-CEDN} must learn to better leverage reciprocal Gabor filters to fill holes caused by the lack of acoustic returns in certain target facets.  It must also construct complex, shape-based filters for faint target boundaries.  Nevertheless, much of the target is correctly deemed visually salient.  Its overall quality is on par with the sonar-adapted {\sc DHS} and {\sc MCDL} networks.

Taken together, our findings indicate that pre-training the {\sc MB-CEDN}s on natural-image datasets leads to initial contrast filters that segment multi-aspect sonar imagery well. 

\setstretch{0.95}\fontsize{9.75}{10}\selectfont
\putbib
\end{bibunit}

\end{document}